\documentclass[10pt,twocolumn,letterpaper]{article}

\usepackage{iccv}              

\definecolor{iccvblue}{rgb}{0.21,0.49,0.74}
\usepackage[pagebackref,breaklinks,colorlinks,allcolors=iccvblue]
{hyperref}

\usepackage{times}
\usepackage{epsfig}
\usepackage{graphicx}
\usepackage{amsmath}
\usepackage{amssymb}
\usepackage{tabularx}
\usepackage[pagebackref]{hyperref}
\usepackage{import}
\usepackage{stfloats}
\usepackage{verbatim}
\usepackage{xcolor,colortbl}
\usepackage[section]{placeins}
\usepackage{subcaption}
\usepackage{bm}
\usepackage{soul}
\usepackage{enumitem}
\usepackage{wrapfig}
\usepackage{multirow}
\usepackage{stfloats}       
\usepackage[accsupp]{axessibility}

\def\paperID{9} 
\def\confName{ICCV}
\def\confYear{2025}

\title{ViT-FIQA: Assessing Face Image Quality using Vision Transformers}

\author{
Andrea Atzori\textsuperscript{1} \quad
Fadi Boutros\textsuperscript{1} \quad
Naser Damer\textsuperscript{1,2} \\
\small \textsuperscript{1}Fraunhofer Institute for Computer Graphics Research IGD, Darmstadt, Germany \\
\small \textsuperscript{2}Department of Computer Science, TU Darmstadt, Germany \\
{\tt\small andrea.atzori@igd-extern.fraunhofer.de, \{fadi.boutros, naser.damer\}@igd.fraunhofer.de}
}

\begin{document}
\maketitle
\begin{abstract}

Face Image Quality Assessment (FIQA) aims to predict the utility of a face image for face recognition (FR) systems. State-of-the-art FIQA methods mainly rely on convolutional neural networks (CNNs), leaving the potential of Vision Transformer (ViT) architectures underexplored. This work proposes ViT-FIQA, a novel approach that extends standard ViT backbones, originally optimized for FR, through a learnable quality token designed to predict a scalar utility score for any given face image. The learnable quality token is concatenated with the standard image patch tokens, and the whole sequence is processed via global self-attention by the ViT encoders to aggregate contextual information across all patches. At the output of the backbone, ViT-FIQA branches into two heads: (1) the patch tokens are passed through a fully connected layer to learn discriminative face representations via a margin-penalty softmax loss, and (2) the quality token is fed into a regression head to learn to predict the face sample's utility. Extensive experiments on challenging benchmarks and several FR models, including both CNN- and ViT-based architectures, demonstrate that ViT-FIQA consistently achieves top-tier performance. These results underscore the effectiveness of transformer-based architectures in modeling face image utility and highlight the potential of ViTs as a scalable foundation for future FIQA research \url{https://cutt.ly/irHlzXUC}.

\end{abstract}    
\vspace{-2mm}
\section{Introduction}
\label{sec:intro}
\vspace{-1mm}

FIQA estimates the utility value of a face sample for FR \cite{ISOIEC29794-1}. These algorithms analyze a face image to generate a scalar quality score - namely the Face Image Quality (FIQ) score - to quantitatively measure the face image's suitability for use in FR systems \cite{NISTQuaity}. High FIQA scores suggest better sample recognisability, thus increasing FR efficiency in applications such as automated border control \cite{NISTQuaity, DBLP:journals/csur/SchlettRHGFB22}. FIQA focuses on evaluating the utility of a face image for automated FR \cite{NISTQuaity, ISOIEC29794-1, DBLP:journals/csur/SchlettRHGFB22}, and does not aim to measure (nor reflect) the perceived image quality. The perceived image quality has been explored through general image quality assessment (IQA) methods \cite{liu2017rankiqa, BRISQE_IQA, mittal2012making}, which offer insights into image quality from the perspective of human perception. However, IQA does not necessarily correlate with the face image's utility for FR \cite{BiyingWACV}. For example, a face image might score high on perceived quality according to IQA metrics but still be less suitable for FR due to issues like occlusion or extreme pose \cite{terhorst2020face}. Conversely, FIQAs provide a focused evaluation of the image's utility for FR tasks. Thus, as noted in \cite{SDDFIQA, DBLP:conf/iwbf/BabnikDS23, boutros_2023_crfiqa, MagFace}, FIQA methods in literature demonstrate superiority over IQA when assessing the utility of a face image for FR.

This paper proposes ViT-FIQA, a novel approach that extends ViTs to predict FIQ. We achieved that by extending the standard ViT designed for FR with a dedicated learnable token that attends to all image patches and learns to regress a scalar quality score. The quality token is processed as an image token with a position encoding of zero and initialized as a learnable embedding. Then, it is added to the sequence of patch embeddings at the input of the transformer. At each transformer layer, it attends to all the image patches via self-attention, gathering global, contextual information about the whole face image. This final representation of the quality token is passed through a regression layer to output a scalar quality score. This regression layer is optimized using CR-FIQA loss \cite{boutros_2023_crfiqa}, which quantifies the relative classifiability of a sample as the ratio between the similarity to its class center and that of its closest negative class. This integration ensures that the predicted quality score aligns with the underlying FR task, maintaining tight coupling between the quality assessment and the recognition pipeline. 
We conducted extensive evaluations of ViT-FIQA across diverse benchmarks (LFW, CA-LFW, AgeDB-30, Adience, XQLFW, CFP-FP, CP-LFW), under cross-model FR evaluation setups using both CNN-based and ViT-based FR models. Our results show that ViT-FIQA consistently ranks among top-performing methods, particularly excelling in pose- and quality-challenging datasets. By exploring the synergy between ViTs and FIQA, our work offers new insights into using ViT for modeling face image utility, advancing the SOTA  in several experimental setups.

\section{Related Work}


\paragraph{ViTs in Face Recognition:}
ViTs \cite{DBLP:conf/iclr/DosovitskiyB0WZ21} have emerged as powerful competitors to CNNs, achieving comparable performance on various vision tasks, including segmentation \cite{DBLP:conf/iccv/KirillovMRMRGXW23} and detection \cite{DBLP:conf/iclr/0097LL000NS23} despite lacking convolutional inductive biases. Early attempts to apply ViTs to FR demonstrated the viability of transformer-based architectures in this domain. FaceTransformer \cite{Zhong_2021_FT} adopted pure-transformer backbones and achieved competitive accuracy on verification benchmarks. ClusFormer \cite{Nguyen_2021_CVPR} introduced an unsupervised transformer-driven approach for clustering faces at a large scale, outperforming previous unsupervised clustering methods in FR tasks. 
Part-fViT \cite{DBLP:conf/bmvc/SunT22} introduced a ViT-based FR pipeline in two stages. First, a fViT backbone is proposed as a baseline, trained with a standard margin-based loss. Second, a part-based ViT is used to learn to predict facial landmark locations without the need for ground-truths, in order to feed the patches extracted in the predicted locations to the fViT.
TransFace \cite{Dan_2023_TransFace} introduced DPAP (Dominant Patch Amplitude Perturbation), a patch-level augmentation that perturbs dominant patch pixels to increase data diversity without losing facial structure, and EHSM (Entropy-based Hard Sample Mining), which uses token entropy to weight easy vs. hard samples during training.
S-ViT \cite{Kim_2023_SViT} proposed a sparse vision transformer pruned for FR efficiency, which prunes redundant attention heads and tokens during training to reduce model size without sacrificing accuracy.
KP-RPE \cite{DBLP:conf/cvpr/KimS0JL24} introduced a method to make ViT models more robust to misalignment and pose variations in face images by leveraging facial landmarks in the positional encodings. Since Relative Position Encoding (RPE) biases ViT attention toward nearby patches, the proposed KP-RPE anchors the positional importance relative to landmark locations (e.g. eyes, nose) rather than just patch distance, achieving marked improvements in accuracy when face alignment fails or faces undergo affine transformations, enhancing robustness to occlusions and pose changes.

\paragraph{Face Image Quality Assessment:}
Existing FIQA approaches fall into four groups. 
(1) Label-generation approaches \cite{faceqnetv1,SDDFIQA,best2018learning} train regression networks using quality labels from human assessments \cite{best2018learning}, ICAO-compliance comparisons \cite{faceqnetv1}, or distribution distances \cite{SDDFIQA}. RankIQ \cite{RANKIQ_FIQA} adopts a learning-to-rank strategy, predicting quality rankings based on FR performance across datasets to better capture relative differences. These methods often rely on separate, shallow models, limiting their ability to exploit high-capacity FR features. 
(2) Non-FR model approaches include DifFIQA \cite{10449044}, which leverages diffusion models to assess embedding robustness, and eDifFIQA \cite{babnikTBIOM2024}, which distills this into a lighter model for faster inference. These approaches can be accurate but are typically computationally expensive due to reliance on large generative models \cite{10449044}. 
(3) Pre-trained FR analysis approaches operate directly on a fixed FR model. SER-FIQ \cite{SERFIQ} measures embedding stability under dropout perturbations, while GraFIQ \cite{grafiqs} uses gradient magnitudes to evaluate alignment with the FR model. FaceQAN \cite{FaceQAN} estimates quality by quantifying adversarial robustness: higher-quality images require larger perturbations to reduce recognition confidence. These methods avoid training but are constrained by the fixed FR backbone. 
(4) FR-integrated approaches embed quality directly into the FR process. MagFace \cite{MagFace} links quality to embedding magnitude via regularized training, while PFE \cite{PFE_FIQA} models embeddings as Gaussians with uncertainty representing quality. CR-FIQA \cite{boutros_2023_crfiqa} estimates quality by predicting a sample's relative classifiability. These methods leverage the full expressiveness of the FR model and achieve SOTA performance on key benchmarks \cite{yang2025fate}.


Despite recent advancements in FIQA, especially in the approaches in the fourth group, and in the utilization of ViTs for FR, the FIQ concept within ViT-based FR was never explored. To address this, we propose a novel FIQA solution that modifies a standard ViT architecture to learn a regression problem to estimate the FIQ score during FR training by (1) using a dedicated trainable token aimed to capture sample utility insights via self-attention and (2) using the original ViT backbone to produce a unique representation for both FR training and FIQ regression, as in \cite{boutros_2023_crfiqa}. 

\vspace{-2mm}
\section{Methodology}
\vspace{-1mm}

\begin{figure*}[!t]
\centering
\includegraphics[width=0.85\textwidth, trim=1 1 1 1,clip]{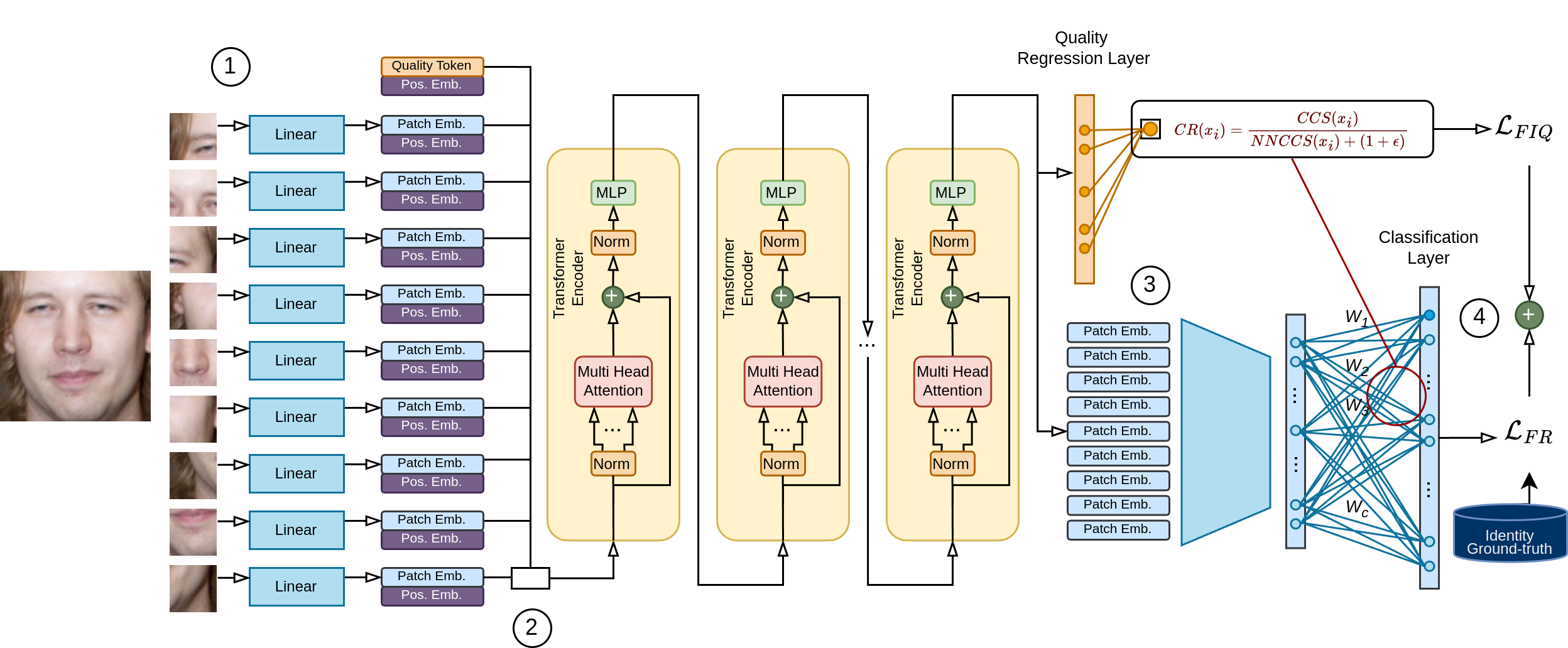}
\vspace{-4mm}
\caption{Overview of our ViT-FIQA. 1) A face sample is divided in equally sized, non-overlapping patches of size $P \times P$. All these patches are then flattened and linearly projected to extract the embeddings. 2) A learnable quality token is concatenated to the patch tokens. The concatenated tokens are fed as input to a sequence of Transformer Encoder layers. 3) The final sequence of embeddings is then used as follows: the first one - the refined quality token - is used as input to the regression layer in order to predict the utility value of the sample, while the remaining patches are used as input for a fully connected layer to obtain a final embedding representing the sample. 4) The two loss terms ($L_{FR}$ and $L_{FIQ}$) are computed and added to obtain the final loss value.}
\label{fig:pipeline}
\vspace{-3mm}
\end{figure*}

This section presents our  FIQA approach that leverages a learnable quality token to capture, through the sequence of attention layers composing the ViTs, discriminant information about the classifiability (utility) of face sample. We provide first details about the ViT architecture followed by our ViT-FIQA.

\paragraph{ViT:}
ViTs~\cite{DBLP:conf/iclr/DosovitskiyB0WZ21} adapt the Transformer architecture, originally designed for NLP, to vision tasks. Unlike CNNs, which exploit local receptive fields and translation-invariant filters, ViTs use global self-attention to model long-range dependencies between image regions. 

\paragraph{Patch Embedding:}
Given an input image $\mathbf{x} \in \mathbb{R}^{H \times W \times C}$, we partition it into $N$ non-overlapping patches of size $P \times P$. Each patch is flattened and linearly projected into a $D$-dimensional embedding space, yielding:
\[
\{\mathbf{e}_1, \mathbf{e}_2, \ldots, \mathbf{e}_N\}, \quad \mathbf{e}_i \in \mathbb{R}^D.
\]
Positional embeddings $\{\mathbf{p}_1, \ldots, \mathbf{p}_N\}$ are added to retain spatial structure:
\[
\mathbf{z}_i^{(0)} = \mathbf{e}_i + \mathbf{p}_i.
\]

\paragraph{Incorporating the Quality Token:}
In ViT-FIQA, we introduce an additional \emph{learnable quality token} $\mathbf{q}_0$ designed to regress a scalar face image utility score. The input sequence becomes:
\[
\mathbf{Z}^{(0)} = [\mathbf{q}_0 + \mathbf{p}_0; \mathbf{z}_1^{(0)}, \ldots, \mathbf{z}_N^{(0)}] \in \mathbb{R}^{(N+1) \times D}.
\]
This sequence is processed by $L$ Transformer encoder layers, each consisting of multi-head self-attention (MSA) and feedforward networks (FFN) with layer normalization and residual connections.

\paragraph{Attention Aggregation of the Quality Token:}
At layer $\ell$, each token $\mathbf{z}_i^{(\ell)}$ is projected into queries, keys, and values:
\[
\mathbf{q}_i^{(\ell)} = \mathbf{z}_i^{(\ell)} W_Q, 
\quad \mathbf{k}_i^{(\ell)} = \mathbf{z}_i^{(\ell)} W_K, 
\quad \mathbf{v}_i^{(\ell)} = \mathbf{z}_i^{(\ell)} W_V,
\]

with $W_Q, W_K, W_V \in \mathbb{R}^{D \times d_h}$ and $d_h = D/h$ for $h$ heads. The quality token at position $0$ forms attention weights over all tokens:
\[
\alpha^{(\ell)}_{0,j} = 
\frac{\exp \left( \tfrac{\mathbf{q}_0^{(\ell)} \cdot \mathbf{k}_j^{(\ell)}}{\sqrt{d_h}} \right)}
     {\sum\limits_{m=0}^{N} \exp \left( \tfrac{\mathbf{q}_0^{(\ell)} \cdot \mathbf{k}_m^{(\ell)}}{\sqrt{d_h}} \right)}.
\]
The quality token is updated by forming a linear combination of all value vectors, with coefficients determined by the attention distribution:
\[
\tilde{\mathbf{z}}_0^{(\ell)} = \sum_{j=0}^{N} \alpha^{(\ell)}_{0,j} \, \mathbf{v}_j^{(\ell)}.
\]
This is extended to $h$ parallel heads, whose outputs are then concatenated and linearly projected:
\[
\mathbf{z}_0^{(\ell+1)} = \Big[ \tilde{\mathbf{z}}_0^{(\ell,1)} \, \Vert \, \cdots \, \Vert \, \tilde{\mathbf{z}}_0^{(\ell,h)} \Big] W_O,
\quad W_O \in \mathbb{R}^{(h \cdot d_h) \times D}.
\]

\paragraph{ViT-FIQA:}
Unlike ViT \cite{DBLP:conf/iclr/DosovitskiyB0WZ21}, designed for image classification tasks with a classification token (CLS), the SOTA ViT architecture for FR \cite{Dan_2023_TransFace, Swinface, You_2025_LVFace}, projects the final token embeddings into $D$-dimensional embedding space and pass it through margin-penalty softmax loss, which we extended with quality regression.
We propose to split the final token sequence  into two branches:
\begin{itemize}
    \item The \textbf{quality token} $\mathbf{z}_0^{(L)}$ is used as input for the regression head to predict the scalar quality score $\hat{q} \in \mathbb{R}$, which is optimized through CR-FIQA loss \cite{boutros_2023_crfiqa}.
    \item The \textbf{patch tokens} $\{\mathbf{z}_i^{(L)}\}_{i=1}^{N}$ are projected into a $D$-dimensional embedding using a fully-connected layer and used for the FR task, trained with  CosFace~\cite{DBLP:conf/cvpr/WangWZJGZL018}.
\end{itemize}
The final loss to jointly learn FR and quality prediction is given by:
\[
\mathcal{L} = \mathcal{L}_{\text{FR}} + \lambda \mathcal{L}_{\text{FIQ}},
\]
Where $\mathcal{L}_{\text{FR}}$ is the FR loss (CosFace~\cite{DBLP:conf/cvpr/WangWZJGZL018} loss),
$\mathcal{L}_{\text{FIQ}}$ is the FIQA regression loss (Smooth L1 between predicted quality and CR-FIQ \cite{boutros_2023_crfiqa} score), and $\lambda$ is a weighting factor (set to 10, following \cite{boutros_2023_crfiqa}) to balance the two objectives. 
By integrating the quality token directly into the Transformer pipeline, ViT-FIQA leverages the full capacity of ViTs to model face image utility in a principled, scalable manner.

\vspace{-2mm}
\section{Experimental Setup}
\vspace{-1mm}

\paragraph{Datasets:}
We utilized MS1MV2 \cite{deng2019arcface} to train our ViT-FIQ. MS1MV2 \cite{deng2019arcface} is a refined version of  MS-Celeb-1M~\cite{guo_2016_ms1m} dataset. It contains approximately 5.8M images from 85K identities. All images are aligned and cropped to $112 \times 112$ using landmarks obtained by the Multi-task Cascaded Convolutional Networks (MTCNN)~\cite{zhang2016joint}, following~\cite{deng2019arcface}. All the training and testing images are normalized to have values between -1 and 1.

\paragraph{Model Training:}
We utilized ViT-Small \cite{DBLP:conf/iclr/DosovitskiyB0WZ21} backbone. We trained our ViT-FIQA for 34 epochs using AdamW \cite{DBLP:conf/iclr/LoshchilovH19, DBLP:conf/cvpr/KimS0JL24} optimizer, with mini-batch size of 1024, base learning rate 1e-3 ($\beta_1$ = 0.9, $\beta_2$ = 0.999), weight decay 0.05, and polynomial decay scheduling. We applied an extensive data augmentation pipeline, following \cite{DBLP:conf/cvpr/KimS0JL24}, composed by Affine Transformation (scale, rotation, translation), Cutout / Coarse Dropout, Zero-Padded Random Resized Crop, Brightness Adjustment, Saturation Adjustment, Contrast Adjustment, Sharpness Adjustment, Histogram Equalization, Grayscale Conversion, Average Blur, Gaussian Blur, Motion Blur, Low-Resolution Simulation via Resize, Spatial Warping. We compared the effectiveness of the learnable quality token with respect to the original method proposed by \cite{boutros_2023_crfiqa}, we trained our models with both methods. The former, which employs the novel quality token, will be noted as ViT-FIQA (T). In the latter, noted as ViT-FIQA (C), no additional token has been employed, leaving the final face embedding as input for both the margin-penalty softmax loss and the FIQ regression layer.


\paragraph{Evaluation:}
We reported the achieved results on seven different benchmarks:  Labeled Faces in the Wild (LFW) \cite{LFWTech}, AgeDB-30 \cite{agedb}, Celebrities in Frontal-Profile in the Wild (CFP-FP) \cite{cfp-fp}, Cross-Age LFW (CALFW) \cite{CALFW}, Adience \cite{Adience}, Cross-Pose LFW (CPLFW) \cite{CPLFWTech}, and Cross-Quality LFW (XQLFW) \cite{XQLFW}. These benchmarks enable a comparative analysis with state-of-the-art (SOTA) FIQA techniques \cite{boutros_2023_crfiqa,MagFace,DBLP:conf/iwbf/BabnikDS23,babnikTBIOM2024} and allow us to assess the robustness of the proposed approach. Performance is assessed via Error-versus-Discard Characteristic (EDC) curves \cite{GT07}, which evaluate the effect of excluding low-quality facial images on verification accuracy. The False Non-Match Rate (FNMR) is measured at set False Match Rate (FMR) thresholds \cite{iso_metric}, notably $1e-3$, as advised for border security by Frontex \cite{frontex2015best}, and $1e-4$. Furthermore, the Area Under the Curve (AUC) as well as the partial Area Under the Curve (pAUC) of the EDC are reported to assess performance across different rejection rates. The pAUC assesses verification effectiveness by focusing on a specific section of the EDC, up to a rejection rate of 30\%, according to \cite{10449044, babnikTBIOM2024, Ou_2024_CVPR, DBLP:journals/tbbis/SchlettRTB24}. To evaluate the influence of FIQA on varied FR models, we conduct experiments with four CNN-based FR systems: ArcFace \cite{deng2019arcface}, ElasticFace (ElasticFace-Arc) \cite{elasticface}, MagFace \cite{MagFace}, and CurricularFace \cite{curricularFace} and two ViT-based FR: SwinFace \cite{Swinface} and TransFace \cite{Dan_2023_TransFace}, following \cite{babnikTBIOM2024}.

\vspace{-2mm}
\section{Experimental Results}
\begin{figure*}[bh!]
\centering
\includegraphics[width=0.95\textwidth, trim=1 1 1 1,clip]{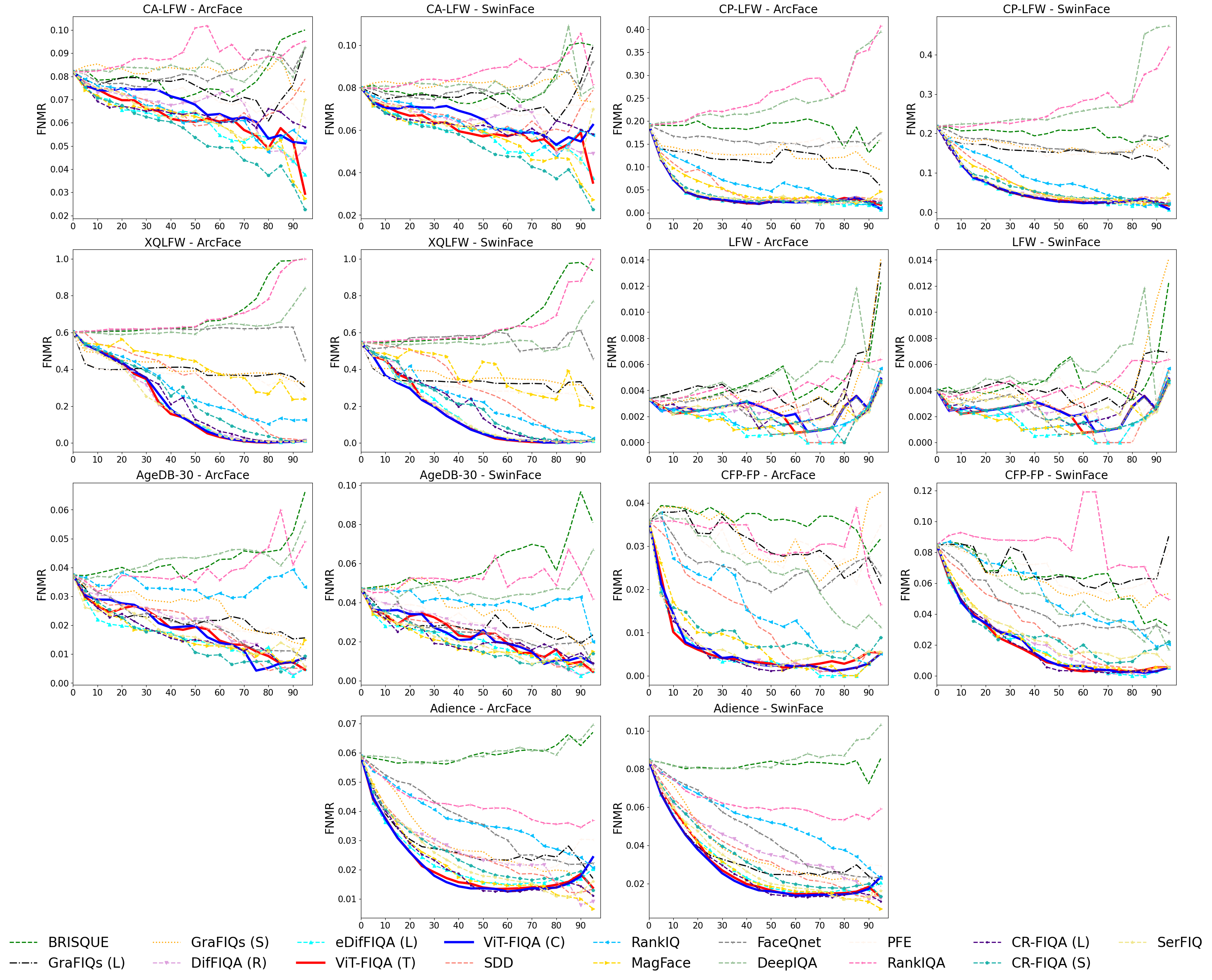}
\vspace{-2mm}
\caption{EDC curves for FNMR@FMR=1e-3 for all evaluated benchmarks using CNN-based (ArcFace) and ViT-based (SwinFace) FR models. The proposed ViT-FIQA (T) is shown in \textcolor{red}{solid red}, ViT-FIQA (C) is shown in \textcolor{blue}{solid blue}.}
\label{fig:fiqa_cnn}
\vspace{-3mm}
\end{figure*}

\vspace{-1mm}
This section provides empirical proof of the validity of the use of the quality token together with ViTs to estimate the utility of face samples. Table \ref{tab:auc_all} presents the verification performance as AUC at FMR=$1e-3$ and FMR=$1e-4$, reported on seven benchmarks while experimenting with four FR solutions, Table \ref{tab:pauc_all} as partial AUC (30\%) at FMR=$1e-3$ and FMR=$1e-4$ reported on the same settings. The EDCs are provided in Figure \ref{fig:fiqa_cnn}. We compared our results with three IQA approaches, BRISQUE, RankIQA, and DeepIQA, as well as with twelve SOTA FIQA approaches, RankIQ, PFE, SER-FIQ, FaceQnet, MagFace, SDD-FIQA, and CR-FIQA (S and L), DifFIQA, eDifFIQA, and GraFIQs (S and L).

\begin{table*}[h!]
 \begin{center}
 
 \resizebox{.95\textwidth}{!}{
\begin{tabular}{|c|cl|cc|cc|cc|cc|cc|cc|cc|cc|}
 \hline
    \multirow{2}{*}{FR} & \multicolumn{2}{c|}{\multirow{2}{*}{Method}} & \multicolumn{2}{c|}{Adience\cite{Adience}} & \multicolumn{2}{c|}{AgeDB-30\cite{agedb}} & \multicolumn{2}{c|}{CFP-FP\cite{cfp-fp}} & \multicolumn{2}{c|}{LFW\cite{LFWTech}} & \multicolumn{2}{c|}{CALFW\cite{CALFW}} & \multicolumn{2}{c|}{CPLFW\cite{CPLFWTech}} & \multicolumn{2}{c|}{XQLFW\cite{XQLFW}} & \multicolumn{2}{c|}{Mean AUC} \\ 
                     &             &  & $1e{-3}$ & $1e{-4}$ & $1e{-3}$ & $1e{-4}$ & $1e{-3}$ & $1e{-4}$ & $1e{-3}$ & $1e{-4}$ & $1e{-3}$ & $1e{-4}$ & $1e{-3}$ & $1e{-4}$ & $1e{-3}$ & $1e{-4}$ & $1e{-3}$ & $1e{-4}$ \\  \hline
  \hline 
\multirow{17}{*}{\rotatebox[origin=c]{90}{ArcFace}} &  \multirow{3}{*}{\rotatebox[origin=c]{90}{IQA}} & BRISQUE\cite{BRISQE_IQA} & 0.0565 & 0.1289 & 0.0400 & 0.0585 & 0.0344 & 0.0433 & 0.0043 & 0.0049 & 0.0755 & 0.0813 & 0.1755 & 0.2166 & 0.6679 & 0.7122 & 0.0644 & 0.0889 \\
 &  & RankIQA\cite{liu2017rankiqa} & 0.0400 & 0.0933 & 0.0372 & 0.0523 & 0.0301 & 0.0384 & 0.0039 & 0.0045 & 0.0846 & 0.0915 & 0.2437 & 0.2969 & 0.6584 & 0.7039 & 0.0733 & 0.0962 \\
 &  & DeepIQA\cite{DEEPIQ_IQA} & 0.0568 & 0.1372 & 0.0403 & 0.0523 & 0.0238 & 0.0292 & 0.0049 & 0.0056 & 0.0793 & 0.0850 & 0.2309 & 0.2856 & 0.5958 & 0.6458 & 0.0727 & 0.0991 \\
\cline{2-19}
 &  \multirow{14}{*}{\rotatebox[origin=c]{90}{FIQA}} & RankIQ\cite{RANKIQ_FIQA} & 0.0353 & 0.0873 & 0.0322 & 0.0420 & 0.0152 & 0.0260 & 0.0018 (3) & 0.0024 (3) & 0.0608 & 0.0672 & 0.0633 & 0.0848 & 0.2789 & 0.3332 & 0.0348 & 0.0516 \\
 &  & PFE\cite{PFE_FIQA} & 0.0273 & 0.0483 & 0.0218 & 0.0289 & 0.0287 & 0.0347 & 0.0030 & 0.0038 & 0.0682 & 0.0718 & 0.1343 & 0.1605 & 0.3569 & 0.4187 & 0.0472 & 0.0580 \\
 &  & SER-FIQ\cite{SERFIQ} & 0.0223 & 0.0434 & 0.0167 & 0.0223 & 0.0065 & 0.0103 & 0.0023 & 0.0028 & 0.0595 & 0.0627 (3) & 0.0389 & 0.0584 & \textbf{0.1812* (1)} & 0.2295* (2) & 0.0244 & 0.0333 \\
 &  & FaceQnet\cite{hernandez2019faceqnet,faceqnetv1} & 0.0346 & 0.0734 & 0.0197 & 0.0245 & 0.0240 & 0.0273 & 0.0022 & 0.0027 & 0.0774 & 0.0822 & 0.1504 & 0.1751 & 0.5829 & 0.6136 & 0.0514 & 0.0642 \\
 &  & MagFace\cite{MagFace} & 0.0207 & 0.0425 & 0.0156 (3) & 0.0198 & 0.0073 & 0.0105 & 0.0016 (2) & 0.0021 (2) & 0.0568 (2) & 0.0602 (2) & 0.0492 & 0.0642 & 0.4022 & 0.4636 & 0.0252 & 0.0332 \\
 &  & SDD-FIQA\cite{SDDFIQA} & 0.0248 & 0.0562 & 0.0186 & 0.0206 & 0.0122 & 0.0193 & 0.0021 & 0.0027 & 0.0641 & 0.0698 & 0.0517 & 0.0670 & 0.3090 & 0.3561 & 0.0289 & 0.0393 \\
 &  & CR-FIQA(S) \cite{boutros_2023_crfiqa} & 0.0241 & 0.0517 & \textbf{0.0144 (1)} & 0.0187 (2) & 0.0090 & 0.0145 & 0.0020 & 0.0025 & \textbf{0.0521 (1)} & \textbf{0.0554 (1)} & 0.0391 & 0.0567 & 0.2377 & 0.2740 & 0.0234 (2) & 0.0333 \\
 &  & CR-FIQA(L) \cite{boutros_2023_crfiqa} & 0.0204 (3) & \textbf{0.0353 (1)} & 0.0159 & 0.0189 (3) & 0.0050 (2) & 0.0082 (2) & 0.0023 & 0.0029 & 0.0616 & 0.0632 & 0.0360 (2) & 0.0515 (2) & 0.2084 & 0.2441 & 0.0235 & \textbf{0.0300 (1)} \\
 &  & DifFIQA(R) \cite{10449044} & 0.0251 & 0.0619 & 0.0194 & 0.0262 & 0.0053 (3) & 0.0091 & 0.0020 & 0.0025 & 0.0629 & 0.0688 & 0.0365 & 0.0531 & 0.1847 (2) & 0.2397 (3) & 0.0252 & 0.0369 \\
 &  & eDifFIQA(L) \cite{babnikTBIOM2024} & 0.0210 & 0.0402 & 0.0148 (2) & \textbf{0.0176 (1)} & \textbf{0.0049 (1)} & 0.0083 (3) & \textbf{0.0014 (1)} & \textbf{0.0019 (1)} & 0.0574 (3) & 0.0627 & \textbf{0.0342 (1)} & \textbf{0.0500 (1)} & 0.1917 & 0.2469 & \textbf{0.0223 (1)} & 0.0301 (2) \\
 &  & GraFIQs (S) \cite{grafiqs} & 0.0279 & 0.0517 & 0.0247 & 0.0399 & 0.0305 & 0.0398 & 0.0034 & 0.0041 & 0.0787 & 0.0842 & 0.1228 & 0.1573 & 0.3909 & 0.4263 & 0.0480 & 0.0628 \\
 &  & GraFIQs (L) \cite{grafiqs} & 0.0269 & 0.0429 & 0.0211 & 0.0306 & 0.0292 & 0.0389 & 0.0041 & 0.0047 & 0.0714 & 0.0759 & 0.1120 & 0.1417 & 0.3731 & 0.4234 & 0.0441 & 0.0558 \\
 &  & ViT-FIQA (T) (Ours) & 0.0197 (2) & 0.0395 (3) & 0.0177 & 0.0207 & 0.0057 & 0.0084 & 0.0023 & 0.0027 & 0.0593 & 0.0627 & 0.0366 & 0.0519 (3) & 0.1864 (3) & \textbf{0.2274 (1)} & 0.0235 (3) & 0.0310 (3) \\
 &  & ViT-FIQA (C) (Ours) & \textbf{0.0195 (1)} & 0.0377 (2) & 0.0174 & 0.0225 & 0.0054 & \textbf{0.0081 (1)} & 0.0023 & 0.0028 & 0.0634 & 0.0666 & 0.0364 (3) & 0.0520 & 0.1914 & 0.2637 & 0.0241 & 0.0316 \\
  \hline \hline 
\multirow{17}{*}{\rotatebox[origin=c]{90}{CurricularFace}} &  \multirow{3}{*}{\rotatebox[origin=c]{90}{IQA}} & BRISQUE\cite{BRISQE_IQA} & 0.0502 & 0.1097 & 0.0433 & 0.0491 & 0.0326 & 0.0358 & 0.0041 & 0.0047 & 0.0755 & 0.0784 & 0.1441 & 0.3497 & 0.6146 & 0.6336 & 0.0583 & 0.1046 \\
 &  & RankIQA\cite{liu2017rankiqa} & 0.0359 & 0.0752 & 0.0394 & 0.0510 & 0.0298 & 0.0356 & 0.0039 & 0.0045 & 0.0806 & 0.0865 & 0.2346 & 0.4654 & 0.5900 & 0.6212 & 0.0707 & 0.1197 \\
 &  & DeepIQA\cite{DEEPIQ_IQA} & 0.0492 & 0.1070 & 0.0407 & 0.0476 & 0.0227 & 0.0278 & 0.0050 & 0.0056 & 0.0764 & 0.0786 & 0.2488 & 0.4961 & 0.5165 & 0.5526 & 0.0738 & 0.1271 \\
\cline{2-19}
 &  \multirow{14}{*}{\rotatebox[origin=c]{90}{FIQA}} & RankIQ\cite{RANKIQ_FIQA} & 0.0314 & 0.0715 & 0.0365 & 0.0417 & 0.0186 & 0.0249 & 0.0018 (3) & 0.0024 (3) & 0.0590 & 0.0640 & 0.0541 & 0.0730 & 0.2449 & 0.2880 & 0.0336 & 0.0462 \\
 &  & PFE\cite{PFE_FIQA} & 0.0265 & 0.0430 & 0.0248 & 0.0275 & 0.0314 & 0.0408 & 0.0032 & 0.0038 & 0.0678 & 0.0705 & 0.1160 & 0.1305 & 0.2948 & 0.3171 & 0.0450 & 0.0527 \\
 &  & SER-FIQ\cite{SERFIQ} & 0.0211 & 0.0381 & 0.0167 & \textbf{0.0193 (1)} & 0.0074 & 0.0111 & 0.0025 & 0.0030 & 0.0587 & 0.0610 & 0.0356 & 0.0520 & \textbf{0.1558* (1)} & \textbf{0.1866* (1)} & 0.0237 & 0.0307 \\
 &  & FaceQnet\cite{hernandez2019faceqnet,faceqnetv1} & 0.0326 & 0.0626 & 0.0221 & 0.0267 & 0.0226 & 0.0274 & 0.0022 & 0.0027 & 0.0767 & 0.0799 & 0.1384 & 0.3229 & 0.5035 & 0.5411 & 0.0491 & 0.0870 \\
 &  & MagFace\cite{MagFace} & 0.0200 & 0.0364 & 0.0167 (3) & 0.0195 (3) & 0.0078 & 0.0111 & 0.0016 (2) & 0.0021 (2) & 0.0563 (2) & 0.0590 (2) & 0.0449 & 0.0607 & 0.3758 & 0.4178 & 0.0246 & 0.0315 \\
 &  & SDD-FIQA\cite{SDDFIQA} & 0.0230 & 0.0462 & 0.0219 & 0.0254 & 0.0138 & 0.0185 & 0.0021 & 0.0027 & 0.0637 & 0.0675 & 0.0465 & 0.0671 & 0.2649 & 0.3053 & 0.0285 & 0.0379 \\
 &  & CR-FIQA(S) \cite{boutros_2023_crfiqa} & 0.0227 & 0.0446 & \textbf{0.0156 (1)} & 0.0198 & 0.0097 & 0.0148 & 0.0020 & 0.0025 & \textbf{0.0513 (1)} & \textbf{0.0534 (1)} & 0.0340 & 0.0501 & 0.2101 & 0.2470 & 0.0225 (2) & 0.0309 \\
 &  & CR-FIQA(L) \cite{boutros_2023_crfiqa} & 0.0198 (3) & 0.0336 (2) & 0.0162 (2) & 0.0200 & 0.0054 (2) & \textbf{0.0080 (1)} & 0.0023 & 0.0029 & 0.0605 & 0.0618 & 0.0324 (3) & 0.0462 & 0.1716 & 0.2318 & 0.0228 (3) & 0.0287 (2) \\
 &  & DifFIQA(R) \cite{10449044} & 0.0230 & 0.0475 & 0.0227 & 0.0260 & 0.0055 (3) & 0.0092 & 0.0020 & 0.0025 & 0.0608 & 0.0657 & 0.0305 (2) & \textbf{0.0441 (1)} & 0.1600 (2) & 0.1871 (2) & 0.0241 & 0.0325 \\
 &  & eDifFIQA(L) \cite{babnikTBIOM2024} & 0.0199 & 0.0338 (3) & 0.0170 & 0.0195 (2) & \textbf{0.0048 (1)} & 0.0084 (3) & \textbf{0.0014 (1)} & \textbf{0.0019 (1)} & 0.0566 (3) & 0.0601 (3) & \textbf{0.0303 (1)} & 0.0455 (2) & 0.1717 & 0.2080 & \textbf{0.0217 (1)} & \textbf{0.0282 (1)} \\
 &  & GraFIQs (S) \cite{grafiqs} & 0.0270 & 0.0462 & 0.0271 & 0.0323 & 0.0305 & 0.0392 & 0.0034 & 0.0041 & 0.0768 & 0.0807 & 0.1098 & 0.1259 & 0.3538 & 0.3956 & 0.0458 & 0.0547 \\
 &  & GraFIQs (L) \cite{grafiqs} & 0.0264 & 0.0401 & 0.0209 & 0.0278 & 0.0310 & 0.0417 & 0.0041 & 0.0047 & 0.0696 & 0.0733 & 0.0947 & 0.1110 & 0.3207 & 0.3419 & 0.0411 & 0.0498 \\
 &  & ViT-FIQA (T) (Ours) & 0.0188 (2) & 0.0340 & 0.0198 & 0.0234 & 0.0061 & 0.0086 & 0.0023 & 0.0027 & 0.0587 & 0.0613 & 0.0327 & 0.0461 & 0.1623 (3) & 0.1962 & 0.0231 & 0.0294 (3) \\
 &  & ViT-FIQA (C) (Ours) & \textbf{0.0187 (1)} & \textbf{0.0334 (1)} & 0.0195 & 0.0227 & 0.0058 & 0.0082 (2) & 0.0023 & 0.0028 & 0.0629 & 0.0650 & 0.0326 & 0.0458 (3) & 0.1641 & 0.1962 (3) & 0.0236 & 0.0296 \\
  \hline \hline 
\multirow{17}{*}{\rotatebox[origin=c]{90}{ElasticFace}} &  \multirow{3}{*}{\rotatebox[origin=c]{90}{IQA}} & BRISQUE\cite{BRISQE_IQA} & 0.0643 & 0.1184 & 0.0376 & 0.0403 & 0.0281 & 0.0372 & 0.0034 & 0.0047 & 0.0725 & 0.0747 & 0.1543 & 0.3196 & 0.6343 & 0.6964 & 0.0600 & 0.0991 \\
 &  & RankIQA\cite{liu2017rankiqa} & 0.0433 & 0.0862 & 0.0374 & 0.0436 & 0.0269 & 0.0318 & 0.0033 & 0.0045 & 0.0810 & 0.0835 & 0.2325 & 0.4306 & 0.6189 & 0.6856 & 0.0707 & 0.1134 \\
 &  & DeepIQA\cite{DEEPIQ_IQA} & 0.0645 & 0.1203 & 0.0384 & 0.0411 & 0.0191 & 0.0256 & 0.0043 & 0.0056 & 0.0756 & 0.0772 & 0.2401 & 0.4541 & 0.5400 & 0.5832 & 0.0737 & 0.1207 \\
\cline{2-19}
 &  \multirow{14}{*}{\rotatebox[origin=c]{90}{FIQA}} & RankIQ\cite{RANKIQ_FIQA} & 0.0400 & 0.0777 & 0.0309 & 0.0337 & 0.0149 & 0.0180 & \textbf{0.0013 (1)} & 0.0020 (2) & 0.0598 & 0.0614 & 0.0581 & 0.0727 & 0.2468 & 0.2776 & 0.0342 & 0.0443 \\
 &  & PFE\cite{PFE_FIQA} & 0.0285 & 0.0461 & 0.0216 & 0.0228 & 0.0258 & 0.0314 & 0.0026 & 0.0038 & 0.0670 & 0.0688 & 0.1225 & 0.1408 & 0.3346 & 0.3544 & 0.0447 & 0.0523 \\
 &  & SER-FIQ\cite{SERFIQ} & 0.0240 & 0.0417 & 0.0163 & 0.0179 & 0.0061 & 0.0085 & 0.0021 & 0.0028 & 0.0574 & 0.0590 & 0.0387 & 0.0513 & \textbf{0.1576* (1)} & \textbf{0.1868* (1)} & 0.0241 & 0.0302 \\
 &  & FaceQnet\cite{hernandez2019faceqnet,faceqnetv1} & 0.0369 & 0.0667 & 0.0194 & 0.0207 & 0.0227 & 0.0247 & 0.0021 & 0.0026 & 0.0763 & 0.0777 & 0.1420 & 0.2880 & 0.5549 & 0.5844 & 0.0499 & 0.0801 \\
 &  & MagFace\cite{MagFace} & 0.0225 & 0.0385 & 0.0150 & 0.0158 (2) & 0.0069 & 0.0095 & 0.0014 (3) & 0.0021 (3) & 0.0553 (2) & 0.0563 (2) & 0.0474 & 0.0597 & 0.3973 & 0.4282 & 0.0248 & 0.0303 \\
 &  & SDD-FIQA\cite{SDDFIQA} & 0.0277 & 0.0512 & 0.0187 & 0.0200 & 0.0098 & 0.0118 & 0.0019 & 0.0027 & 0.0624 & 0.0638 & 0.0493 & 0.0634 & 0.3052 & 0.3562 & 0.0283 & 0.0355 \\
 &  & CR-FIQA(S) \cite{boutros_2023_crfiqa} & 0.0257 & 0.0465 & 0.0146 (2) & 0.0160 & 0.0070 & 0.0096 & 0.0015 & 0.0022 & \textbf{0.0509 (1)} & \textbf{0.0522 (1)} & 0.0383 & 0.0502 & 0.2093 & 0.2835 & 0.0230 (3) & 0.0294 \\
 &  & CR-FIQA(L) \cite{boutros_2023_crfiqa} & 0.0214 (2) & 0.0357 (2) & 0.0149 (3) & 0.0159 (3) & 0.0045 (2) & \textbf{0.0065 (1)} & 0.0018 & 0.0025 & 0.0594 & 0.0608 & 0.0350 (3) & 0.0462 & 0.1798 & 0.2060 & 0.0228 (2) & 0.0279 (2) \\
 &  & DifFIQA(R) \cite{10449044} & 0.0278 & 0.0536 & 0.0194 & 0.0207 & 0.0050 & 0.0073 & 0.0019 & 0.0025 & 0.0616 & 0.0634 & 0.0330 (2) & 0.0445 (2) & 0.1599 (2) & 0.1890 (2) & 0.0248 & 0.0320 \\
 &  & eDifFIQA(L) \cite{babnikTBIOM2024} & 0.0222 & 0.0374 & \textbf{0.0139 (1)} & \textbf{0.0148 (1)} & \textbf{0.0043 (1)} & 0.0066 (2) & 0.0014 (2) & \textbf{0.0019 (1)} & 0.0564 (3) & 0.0576 (3) & \textbf{0.0323 (1)} & \textbf{0.0440 (1)} & 0.1688 & 0.1996 (3) & \textbf{0.0217 (1)} & \textbf{0.0270 (1)} \\
 &  & GraFIQs (S) \cite{grafiqs} & 0.0297 & 0.0512 & 0.0265 & 0.0303 & 0.0236 & 0.0322 & 0.0031 & 0.0041 & 0.0773 & 0.0791 & 0.1209 & 0.1395 & 0.3854 & 0.4414 & 0.0468 & 0.0561 \\
 &  & GraFIQs (L) \cite{grafiqs} & 0.0280 & 0.0430 & 0.0201 & 0.0224 & 0.0245 & 0.0292 & 0.0037 & 0.0046 & 0.0695 & 0.0711 & 0.1035 & 0.1231 & 0.3598 & 0.3963 & 0.0416 & 0.0489 \\
 &  & ViT-FIQA (T) (Ours) & 0.0214 (3) & 0.0362 (3) & 0.0169 & 0.0179 & 0.0052 & 0.0073 & 0.0017 & 0.0023 & 0.0575 & 0.0592 & 0.0354 & 0.0461 & 0.1698 & 0.2174 & 0.0230 & 0.0282 (3) \\
 &  & ViT-FIQA (C) (Ours) & \textbf{0.0211 (1)} & \textbf{0.0353 (1)} & 0.0167 & 0.0176 & 0.0046 (3) & 0.0066 (3) & 0.0017 & 0.0023 & 0.0613 & 0.0628 & 0.0352 & 0.0459 (3) & 0.1659 (3) & 0.2133 & 0.0234 & 0.0284 \\
  \hline \hline 
\multirow{17}{*}{\rotatebox[origin=c]{90}{MagFace}} &  \multirow{3}{*}{\rotatebox[origin=c]{90}{IQA}} & BRISQUE\cite{BRISQE_IQA} & 0.0594 & 0.1309 & 0.0442 & 0.0799 & 0.0422 & 0.0589 & 0.0043 & 0.0058 & 0.0758 & 0.0788 & 0.2916 & 0.5097 & 0.6911 & 0.7229 & 0.0863 & 0.1440 \\
 &  & RankIQA\cite{liu2017rankiqa} & 0.0407 & 0.0889 & 0.0370 & 0.0681 & 0.0369 & 0.0543 & 0.0041 & 0.0056 & 0.0829 & 0.0857 & 0.3251 & 0.6475 & 0.6706 & 0.7046 & 0.0878 & 0.1583 \\
 &  & DeepIQA\cite{DEEPIQ_IQA} & 0.0571 & 0.1302 & 0.0417 & 0.0721 & 0.0322 & 0.0545 & 0.0048 & 0.0059 & 0.0787 & 0.0809 & 0.3672 & 0.6632 & 0.6162 & 0.6519 & 0.0969 & 0.1678 \\
\cline{2-19}
 &  \multirow{14}{*}{\rotatebox[origin=c]{90}{FIQA}} & RankIQ\cite{RANKIQ_FIQA} & 0.0359 & 0.0837 & 0.0361 & 0.0531 & 0.0213 & 0.0332 & 0.0019 (3) & 0.0027 & 0.0602 & 0.0629 & 0.0659 & 0.1642 & 0.3076 & 0.3475 & 0.0369 & 0.0666 \\
 &  & PFE\cite{PFE_FIQA} & 0.0283 & 0.0481 & 0.0244 & 0.0401 & 0.0398 & 0.0625 & 0.0029 & 0.0035 & 0.0680 & 0.0691 & 0.1357 & 0.1815 & 0.4089 & 0.4472 & 0.0499 & 0.0675 \\
 &  & SER-FIQ\cite{SERFIQ} & 0.0233 & 0.0451 & 0.0185 & 0.0293 & 0.0080 & 0.0139 & 0.0025 & 0.0033 & 0.0590 & 0.0607 & 0.0397 & 0.0821 & \textbf{0.2139* (1)} & \textbf{0.2562* (1)} & 0.0252 & 0.0391 \\
 &  & FaceQnet\cite{hernandez2019faceqnet,faceqnetv1} & 0.0365 & 0.0720 & 0.0217 & 0.0314 & 0.0271 & 0.0351 & 0.0022 & 0.0027 & 0.0763 & 0.0773 & 0.2988 & 0.5218 & 0.6016 & 0.6210 & 0.0771 & 0.1234 \\
 &  & MagFace\cite{MagFace} & 0.0212 & 0.0417 & \textbf{0.0159 (1)} & 0.0247 & 0.0085 & 0.0129 & 0.0017 (2) & 0.0022 (2) & 0.0562 (2) & 0.0578 (2) & 0.0506 & 0.0887 & 0.4478 & 0.4900 & 0.0257 & 0.0380 \\
 &  & SDD-FIQA\cite{SDDFIQA} & 0.0253 & 0.0562 & 0.0216 & 0.0305 & 0.0146 & 0.0201 & 0.0021 & 0.0027 & 0.0643 & 0.0657 & 0.0525 & 0.1188 & 0.3404 & 0.3928 & 0.0301 & 0.0490 \\
 &  & CR-FIQA(S) \cite{boutros_2023_crfiqa} & 0.0244 & 0.0507 & 0.0165 (2) & \textbf{0.0234 (1)} & 0.0102 & 0.0121 & 0.0020 & 0.0028 & \textbf{0.0516 (1)} & \textbf{0.0528 (1)} & 0.0409 & 0.0840 & 0.2670 & 0.3336 & 0.0243 (3) & 0.0376 \\
 &  & CR-FIQA(L) \cite{boutros_2023_crfiqa} & 0.0211 (3) & \textbf{0.0372 (1)} & 0.0174 & 0.0235 (2) & 0.0062 (2) & \textbf{0.0080 (1)} & 0.0023 & 0.0028 & 0.0614 & 0.0628 & 0.0374 (3) & 0.0679 (2) & 0.2369 & 0.2839 & 0.0243 & \textbf{0.0337 (1)} \\
 &  & DifFIQA(R) \cite{10449044} & 0.0256 & 0.0585 & 0.0223 & 0.0363 & 0.0066 & 0.0150 & 0.0020 & 0.0025 (3) & 0.0638 & 0.0660 & 0.0371 (2) & 0.0851 & 0.2177 (2) & 0.2642 (2) & 0.0262 & 0.0439 \\
 &  & eDifFIQA(L) \cite{babnikTBIOM2024} & 0.0216 & 0.0403 & 0.0168 (3) & 0.0246 (3) & \textbf{0.0058 (1)} & 0.0121 & \textbf{0.0014 (1)} & \textbf{0.0019 (1)} & 0.0580 (3) & 0.0595 (3) & \textbf{0.0357 (1)} & 0.0810 & 0.2278 & 0.2792 & \textbf{0.0232 (1)} & 0.0366 \\
 &  & GraFIQs (S) \cite{grafiqs} & 0.0297 & 0.0550 & 0.0282 & 0.0582 & 0.0351 & 0.0589 & 0.0034 & 0.0044 & 0.0777 & 0.0819 & 0.1283 & 0.1832 & 0.4329 & 0.4993 & 0.0504 & 0.0736 \\
 &  & GraFIQs (L) \cite{grafiqs} & 0.0284 & 0.0456 & 0.0215 & 0.0451 & 0.0404 & 0.0482 & 0.0041 & 0.0050 & 0.0704 & 0.0722 & 0.1155 & 0.1679 & 0.4274 & 0.4628 & 0.0467 & 0.0640 \\
 &  & ViT-FIQA (T) (Ours) & 0.0201 (2) & 0.0397 (3) & 0.0186 & 0.0276 & 0.0069 & 0.0099 (2) & 0.0023 & 0.0028 & 0.0590 & 0.0604 & 0.0377 & \textbf{0.0678 (1)} & 0.2211 & 0.2692 (3) & 0.0241 (2) & 0.0347 (2) \\
 &  & ViT-FIQA (C) (Ours) & \textbf{0.0197 (1)} & 0.0381 (2) & 0.0186 & 0.0295 & 0.0064 (3) & 0.0114 (3) & 0.0024 & 0.0028 & 0.0634 & 0.0648 & 0.0375 & 0.0684 (3) & 0.2187 (3) & 0.2706 & 0.0247 & 0.0358 (3) \\
  \hline \hline 
\multirow{17}{*}{\rotatebox[origin=c]{90}{SwinFace}} &  \multirow{3}{*}{\rotatebox[origin=c]{90}{IQA}} & BRISQUE\cite{BRISQE_IQA} & 0.0778 & 0.1833 & 0.0568 & 0.0627 & 0.0605 & 0.1508 & 0.0045 & 0.0059 & 0.0760 & 0.0865 & 0.1912 & 0.2779 & 0.6211 & 0.6707 & 0.0778 & 0.1278 \\
 &  & RankIQA\cite{liu2017rankiqa} & 0.0587 & 0.1409 & 0.0496 & 0.0575 & 0.0809 & 0.2503 & 0.0043 & 0.0058 & 0.0830 & 0.0905 & 0.2519 & 0.3661 & 0.6001 & 0.6581 & 0.0881 & 0.1519 \\
 &  & DeepIQA\cite{DEEPIQ_IQA} & 0.0806 & 0.1766 & 0.0442 & 0.0536 & 0.0489 & 0.1005 & 0.0052 & 0.0062 & 0.0775 & 0.0967 & 0.2584 & 0.3763 & 0.5233 & 0.5892 & 0.0858 & 0.1350 \\
\cline{2-19}
 &  \multirow{14}{*}{\rotatebox[origin=c]{90}{FIQA}} & RankIQ\cite{RANKIQ_FIQA} & 0.0507 & 0.1290 & 0.0391 & 0.0453 & 0.0504 & 0.1552 & 0.0019 (3) & 0.0027 & 0.0603 & 0.0693 & 0.0834 & 0.1071 & 0.2213 & 0.3069 & 0.0476 & 0.0848 \\
 &  & PFE\cite{PFE_FIQA} & 0.0333 & 0.0656 & 0.0269 & 0.0296 & 0.0636 & 0.1678 & 0.0032 & 0.0040 & 0.0679 & 0.0757 & 0.1572 & 0.1865 & 0.2908 & 0.3130 & 0.0587 & 0.0882 \\
 &  & SER-FIQ\cite{SERFIQ} & 0.0289 & 0.0624 & 0.0190 & 0.0228 & 0.0254 & 0.0851 & 0.0027 & 0.0034 & 0.0590 & 0.0614 & 0.0568 & 0.0853 & 0.1453* (3) & 0.1754* (2) & 0.0320 & 0.0534 \\
 &  & FaceQnet\cite{hernandez2019faceqnet,faceqnetv1} & 0.0447 & 0.1014 & 0.0242 & 0.0317 & 0.0419 & 0.1295 & 0.0023 & 0.0028 & 0.0761 & 0.0888 & 0.1652 & 0.2396 & 0.5274 & 0.5870 & 0.0590 & 0.0990 \\
 &  & MagFace\cite{MagFace} & 0.0270 & 0.0601 & \textbf{0.0178 (1)} & \textbf{0.0209 (1)} & 0.0206 & 0.0847 & 0.0017 (2) & 0.0022 (2) & 0.0541 (2) & 0.0593 (2) & 0.0641 & 0.0875 & 0.3627 & 0.4621 & 0.0309 & 0.0524 \\
 &  & SDD-FIQA\cite{SDDFIQA} & 0.0344 & 0.0814 & 0.0216 & 0.0260 & 0.0290 & 0.0881 & 0.0021 & 0.0029 & 0.0627 & 0.0745 & 0.0675 & 0.0937 & 0.2793 & 0.3387 & 0.0362 & 0.0611 \\
 &  & CR-FIQA(S) \cite{boutros_2023_crfiqa} & 0.0322 & 0.0765 & 0.0181 (2) & 0.0222 (2) & 0.0248 & \textbf{0.0581 (1)} & 0.0021 & 0.0029 & \textbf{0.0505 (1)} & \textbf{0.0573 (1)} & 0.0610 & 0.0867 & 0.1893 & 0.2660 & 0.0315 & 0.0506 (2) \\
 &  & CR-FIQA(L) \cite{boutros_2023_crfiqa} & 0.0254 (2) & 0.0557 (2) & 0.0200 & 0.0239 & \textbf{0.0176 (1)} & 0.0854 & 0.0024 & 0.0029 & 0.0599 & 0.0647 & 0.0520 (2) & 0.0749 & 0.1826 & 0.2576 & 0.0295 (2) & 0.0513 \\
 &  & DifFIQA(R) \cite{10449044} & 0.0374 & 0.0937 & 0.0237 & 0.0272 & 0.0201 & 0.0961 & 0.0020 & 0.0026 (3) & 0.0614 & 0.0764 & 0.0521 (3) & 0.0762 & 0.1455 & 0.1865 & 0.0328 & 0.0620 \\
 &  & eDifFIQA(L) \cite{babnikTBIOM2024} & 0.0278 & 0.0589 & 0.0190 (3) & 0.0224 (3) & 0.0180 (3) & 0.0990 & \textbf{0.0014 (1)} & \textbf{0.0019 (1)} & 0.0551 (3) & 0.0658 & \textbf{0.0497 (1)} & \textbf{0.0729 (1)} & 0.1473 & 0.1989 & \textbf{0.0285 (1)} & 0.0535 \\
 &  & GraFIQs (S) \cite{grafiqs} & 0.0353 & 0.0746 & 0.0352 & 0.0443 & 0.0579 & 0.1967 & 0.0041 & 0.0051 & 0.0777 & 0.0898 & 0.1597 & 0.2022 & 0.3436 & 0.3905 & 0.0617 & 0.1021 \\
 &  & GraFIQs (L) \cite{grafiqs} & 0.0322 & 0.0608 & 0.0266 & 0.0359 & 0.0658 & 0.1847 & 0.0039 & 0.0048 & 0.0713 & 0.0801 & 0.1490 & 0.1850 & 0.3190 & 0.3437 & 0.0582 & 0.0919 \\
 &  & ViT-FIQA (T) (Ours) & 0.0264 (3) & 0.0585 (3) & 0.0220 & 0.0264 & 0.0176 (2) & 0.0732 (2) & 0.0023 & 0.0028 & 0.0577 & 0.0607 (3) & 0.0524 & 0.0741 (2) & 0.1402 (2) & 0.1763 (3) & 0.0297 (3) & \textbf{0.0493 (1)} \\
 &  & ViT-FIQA (C) (Ours) & \textbf{0.0253 (1)} & \textbf{0.0556 (1)} & 0.0218 & 0.0256 & 0.0185 & 0.0823 (3) & 0.0024 & 0.0028 & 0.0616 & 0.0654 & 0.0521 & 0.0748 (3) & \textbf{0.1318 (1)} & \textbf{0.1615 (1)} & 0.0303 & 0.0511 (3) \\
  \hline \hline 
\multirow{17}{*}{\rotatebox[origin=c]{90}{TransFace}} &  \multirow{3}{*}{\rotatebox[origin=c]{90}{IQA}} & BRISQUE\cite{BRISQE_IQA} & 0.0487 & 0.0981 & 0.0372 & 0.0429 & 0.0222 & 0.0264 & 0.0032 & 0.0038 & 0.0728 & 0.3662 & 0.1190 & 0.3601 & 0.3461 & 0.3981 & 0.0505 & 0.1496 \\
 &  & RankIQA\cite{liu2017rankiqa} & 0.0357 & 0.0708 & 0.0382 & 0.0442 & 0.0230 & 0.0277 & 0.0028 & 0.0033 & 0.0826 & 0.5354 & 0.1718 & 0.4477 & 0.3359 & 0.3888 & 0.0590 & 0.1882 \\
 &  & DeepIQA\cite{DEEPIQ_IQA} & 0.0476 & 0.0966 & 0.0383 & 0.0441 & 0.0140 & 0.0189 & 0.0035 & 0.0040 & 0.0752 & 0.1767 (2) & 0.2021 & 0.5080 & 0.2847 & 0.3249 & 0.0634 & 0.1414 \\
\cline{2-19}
 &  \multirow{14}{*}{\rotatebox[origin=c]{90}{FIQA}} & RankIQ\cite{RANKIQ_FIQA} & 0.0319 & 0.0612 & 0.0322 & 0.0355 & 0.0117 & 0.0151 & 0.0017 (3) & 0.0022 (3) & 0.0609 & 0.6185 & 0.0527 & 0.0645 & 0.1006 & 0.1445 & 0.0319 & 0.1328 \\
 &  & PFE\cite{PFE_FIQA} & 0.0255 & 0.0391 & 0.0227 & 0.0255 & 0.0219 & 0.0264 & 0.0027 & 0.0033 & 0.0681 & 0.4315 & 0.0971 & 0.1098 & 0.1394 & 0.1679 & 0.0397 & 0.1059 \\
 &  & SER-FIQ\cite{SERFIQ} & 0.0213 & 0.0360 & 0.0156 & 0.0185 & 0.0057 & 0.0082 & 0.0025 & 0.0029 & 0.0575 (3) & 0.3127 (3) & 0.0341 & 0.0456 & 0.0571* & 0.0777* & 0.0228 & 0.0706 (2) \\
 &  & FaceQnet\cite{hernandez2019faceqnet,faceqnetv1} & 0.0319 & 0.0519 & 0.0206 & 0.0235 & 0.0197 & 0.0226 & 0.0018 & 0.0023 & 0.0781 & 0.6981 & 0.1292 & 0.3662 & 0.2331 & 0.3215 & 0.0469 & 0.1941 \\
 &  & MagFace\cite{MagFace} & 0.0201 & 0.0350 & 0.0152 (3) & 0.0166 (3) & 0.0058 & 0.0081 & 0.0016 (2) & 0.0021 (2) & 0.0873 & 0.6961 & 0.0410 & 0.0520 & 0.1647 & 0.1976 & 0.0285 & 0.1350 \\
 &  & SDD-FIQA\cite{SDDFIQA} & 0.0244 & 0.0441 & 0.0191 & 0.0216 & 0.0096 & 0.0127 & 0.0021 & 0.0026 & 0.0967 & 0.7134 & 0.0410 & 0.0527 & 0.1273 & 0.1585 & 0.0321 & 0.1412 \\
 &  & CR-FIQA(S) \cite{boutros_2023_crfiqa} & 0.0233 & 0.0401 & \textbf{0.0138 (1)} & 0.0164 (2) & 0.0059 & 0.0082 & 0.0020 & 0.0026 & \textbf{0.0518 (1)} & 0.6392 & 0.0318 & 0.0432 & 0.0823 & 0.1092 & 0.0214 (2) & 0.1249 \\
 &  & CR-FIQA(L) \cite{boutros_2023_crfiqa} & 0.0197 (2) & 0.0323 (2) & 0.0157 & 0.0174 & 0.0041 & \textbf{0.0056 (1)} & 0.0024 & 0.0028 & 0.0615 & \textbf{0.1217 (1)} & 0.0308 & 0.0422 & 0.0779 & 0.1000 & 0.0224 (3) & \textbf{0.0370 (1)} \\
 &  & DifFIQA(R) \cite{10449044} & 0.0254 & 0.0472 & 0.0181 & 0.0203 & 0.0039 (2) & 0.0068 & 0.0020 & 0.0024 & 0.0618 & 0.5835 & 0.0296 (2) & 0.0409 (2) & 0.0552 & 0.0727 & 0.0235 & 0.1169 \\
 &  & eDifFIQA(L) \cite{babnikTBIOM2024} & 0.0210 & 0.0343 & 0.0143 (2) & \textbf{0.0156 (1)} & \textbf{0.0038 (1)} & 0.0063 (3) & \textbf{0.0014 (1)} & \textbf{0.0018 (1)} & 0.0572 (2) & 0.6371 & \textbf{0.0284 (1)} & \textbf{0.0397 (1)} & 0.0547 (3) & 0.0702 (3) & \textbf{0.0210 (1)} & 0.1225 \\
 &  & GraFIQs (S) \cite{grafiqs} & 0.0267 & 0.0453 & 0.0243 & 0.0281 & 0.0204 & 0.0258 & 0.0034 & 0.0040 & 0.0772 & 0.5322 & 0.0980 & 0.1108 & 0.1759 & 0.2132 & 0.0417 & 0.1244 \\
 &  & GraFIQs (L) \cite{grafiqs} & 0.0262 & 0.0401 & 0.0204 & 0.0252 & 0.0209 & 0.0246 & 0.0036 & 0.0042 & 0.0720 & 0.4442 & 0.0782 & 0.0922 & 0.1752 & 0.2139 & 0.0369 & 0.1051 (3) \\
 &  & ViT-FIQA (T) (Ours) & 0.0197 (3) & 0.0325 (3) & 0.0176 & 0.0194 & 0.0046 & 0.0067 & 0.0022 & 0.0027 & 0.0598 & 0.5664 & 0.0307 (3) & 0.0416 (3) & 0.0490 (2) & 0.0636 (2) & 0.0224 & 0.1116 \\
 &  & ViT-FIQA (C) (Ours) & \textbf{0.0193 (1)} & \textbf{0.0318 (1)} & 0.0175 & 0.0191 & 0.0040 (3) & 0.0060 (2) & 0.0023 & 0.0028 & 0.0638 & 0.6039 & 0.0308 & 0.0420 & \textbf{0.0477 (1)} & \textbf{0.0624 (1)} & 0.0230 & 0.1176 \\
  \hline \hline 
\end{tabular}}
\end{center}
\vspace{-6mm}
\caption{The AUCs of EDC achieved by our ViT-FIQA and the SOTA methods under different FR experimental settings. The notions of $1e-3$ and $1e-4$ indicate the value of the fixed FMR at which the EDC curves (FNMR vs.~reject) were calculated. The results are compared to three IQA and twelve FIQA approaches. The XQLFW dataset uses SER-FIQ (marked with *) as FIQ labeling method.}
\vspace{3mm}
 \label{tab:auc_all}
 \vspace{-6mm}
\end{table*}

\begin{table*}[h!]
 \begin{center}
 
 \resizebox{.95\textwidth}{!}{
\begin{tabular}{|c|cl|cc|cc|cc|cc|cc|cc|cc|cc|}
 \hline
    \multirow{2}{*}{FR} & \multicolumn{2}{c|}{\multirow{2}{*}{Method}} & \multicolumn{2}{c|}{Adience\cite{Adience}} & \multicolumn{2}{c|}{AgeDB-30\cite{agedb}} & \multicolumn{2}{c|}{CFP-FP\cite{cfp-fp}} & \multicolumn{2}{c|}{LFW\cite{LFWTech}} & \multicolumn{2}{c|}{CALFW\cite{CALFW}} & \multicolumn{2}{c|}{CPLFW\cite{CPLFWTech}} & \multicolumn{2}{c|}{XQLFW\cite{XQLFW}} & \multicolumn{2}{c|}{Mean pAUC} \\ 
                     &             &  & $1e{-3}$ & $1e{-4}$ & $1e{-3}$ & $1e{-4}$ & $1e{-3}$ & $1e{-4}$ & $1e{-3}$ & $1e{-4}$ & $1e{-3}$ & $1e{-4}$ & $1e{-3}$ & $1e{-4}$ & $1e{-3}$ & $1e{-4}$ & $1e{-3}$ & $1e{-4}$  \\  \hline
  \hline 
\multirow{17}{*}{\rotatebox[origin=c]{90}{ArcFace}} &  \multirow{3}{*}{\rotatebox[origin=c]{90}{IQA}} & BRISQUE\cite{BRISQE_IQA} & 0.0143 & 0.0333 & 0.0096 & 0.0146 & 0.0096 & 0.0136 & 0.0009 & 0.0010 & 0.0200 & 0.0225 & 0.0480 & 0.0624 & 0.1512 & 0.1689 & 0.0171 & 0.0246 \\
 &  & RankIQA\cite{liu2017rankiqa} & 0.0124 & 0.0302 & 0.0087 & 0.0141 & 0.0088 & 0.0135 & 0.0009 & 0.0010 & 0.0209 & 0.0234 & 0.0506 & 0.0658 & 0.1532 & 0.1709 & 0.0170 & 0.0247 \\
 &  & DeepIQA\cite{DEEPIQ_IQA} & 0.0145 & 0.0337 & 0.0093 & 0.0140 & 0.0088 & 0.0119 & 0.0009 & 0.0010 & 0.0207 & 0.0230 & 0.0504 & 0.0653 & 0.1487 & 0.1668 & 0.0175 & 0.0248 \\
\cline{2-19}
 &  \multirow{14}{*}{\rotatebox[origin=c]{90}{FIQA}} & RankIQ\cite{RANKIQ_FIQA} & 0.0125 & 0.0304 & 0.0090 & 0.0143 & 0.0071 & 0.0114 & 0.0006 & 0.0008 & 0.0191 & 0.0215 & 0.0306 & 0.0427 & 0.1270 & 0.1500 & 0.0132 & 0.0202 \\
 &  & PFE\cite{PFE_FIQA} & 0.0100 & 0.0229 & 0.0080 & 0.0123 & 0.0089 & 0.0124 & 0.0009 & 0.0010 & 0.0192 & 0.0211 & 0.0351 & 0.0470 & \textbf{0.1064 (1)} & 0.1284 (2) & 0.0137 & 0.0195 \\
 &  & SER-FIQ\cite{SERFIQ} & 0.0102 & 0.0244 & 0.0066 & 0.0107 & 0.0035 & 0.0057 & 0.0007 & 0.0008 & 0.0187 & 0.0205 & 0.0199 & 0.0319 & 0.1175* (3) & 0.1385* & 0.0099 & 0.0157 \\
 &  & FaceQnet\cite{hernandez2019faceqnet,faceqnetv1} & 0.0130 & 0.0303 & 0.0076 & 0.0113 & 0.0077 & 0.0100 & 0.0008 & 0.0009 & 0.0196 & 0.0216 & 0.0428 & 0.0554 & 0.1523 & 0.1686 & 0.0153 & 0.0216 \\
 &  & MagFace\cite{MagFace} & 0.0099 & 0.0247 & 0.0065 (2) & 0.0098 & 0.0045 & 0.0068 & \textbf{0.0006 (1)} & 0.0007 (3) & 0.0177 (2) & 0.0193 (2) & 0.0249 & 0.0360 & 0.1359 & 0.1614 & 0.0107 & 0.0162 \\
 &  & SDD-FIQA\cite{SDDFIQA} & 0.0104 & 0.0259 & 0.0073 & 0.0088 (2) & 0.0068 & 0.0109 & 0.0007 & 0.0008 & 0.0188 & 0.0205 & 0.0279 & 0.0377 & 0.1356 & 0.1525 & 0.0120 & 0.0174 \\
 &  & CR-FIQA(S) \cite{boutros_2023_crfiqa} & 0.0101 & 0.0257 & 0.0068 & 0.0100 & 0.0042 & 0.0078 & 0.0006 & 0.0008 & 0.0178 & 0.0197 (3) & 0.0207 & 0.0324 & 0.1242 & 0.1391 & 0.0100 & 0.0161 \\
 &  & CR-FIQA(L) \cite{boutros_2023_crfiqa} & 0.0097 & 0.0201 (2) & 0.0066 (3) & 0.0089 (3) & 0.0035 (3) & 0.0058 & 0.0007 & 0.0009 & \textbf{0.0177 (1)} & \textbf{0.0186 (1)} & 0.0190 (3) & \textbf{0.0307 (1)} & 0.1213 & 0.1378 & 0.0095 (2) & \textbf{0.0142 (1)} \\
 &  & DifFIQA(R) \cite{10449044} & 0.0100 & 0.0256 & 0.0076 & 0.0120 & 0.0036 & 0.0060 & 0.0006 (2) & 0.0007 (2) & 0.0186 & 0.0208 & \textbf{0.0188 (1)} & 0.0308 (3) & 0.1193 & 0.1454 & 0.0099 & 0.0160 \\
 &  & eDifFIQA(L) \cite{babnikTBIOM2024} & 0.0090 (3) & 0.0230 & \textbf{0.0060 (1)} & \textbf{0.0079 (1)} & \textbf{0.0033 (1)} & 0.0056 (3) & 0.0006 (3) & \textbf{0.0007 (1)} & 0.0177 (3) & 0.0199 & 0.0189 (2) & 0.0308 (2) & 0.1223 & 0.1474 & \textbf{0.0093 (1)} & 0.0147 (2) \\
 &  & GraFIQs (S) \cite{grafiqs} & 0.0118 & 0.0262 & 0.0082 & 0.0142 & 0.0095 & 0.0138 & 0.0008 & 0.0009 & 0.0209 & 0.0233 & 0.0361 & 0.0496 & 0.1204 & 0.1354 (3) & 0.0146 & 0.0213 \\
 &  & GraFIQs (L) \cite{grafiqs} & 0.0097 & \textbf{0.0198 (1)} & 0.0069 & 0.0113 & 0.0091 & 0.0135 & 0.0010 & 0.0011 & 0.0196 & 0.0218 & 0.0344 & 0.0472 & 0.1065 (2) & \textbf{0.1282 (1)} & 0.0134 & 0.0191 \\
 &  & ViT-FIQA (T) (Ours) & \textbf{0.0089 (1)} & 0.0231 & 0.0069 & 0.0093 & 0.0033 (2) & \textbf{0.0054 (1)} & 0.0006 & 0.0007 & 0.0184 & 0.0200 & 0.0191 & 0.0309 & 0.1224 & 0.1364 & 0.0096 (3) & 0.0149 (3) \\
 &  & ViT-FIQA (C) (Ours) & 0.0090 (2) & 0.0226 (3) & 0.0074 & 0.0112 & 0.0035 & 0.0055 (2) & 0.0007 & 0.0008 & 0.0189 & 0.0205 & 0.0191 & 0.0311 & 0.1218 & 0.1583 & 0.0098 & 0.0153 \\
  \hline \hline 
\multirow{17}{*}{\rotatebox[origin=c]{90}{CurricularFace}} &  \multirow{3}{*}{\rotatebox[origin=c]{90}{IQA}} & BRISQUE\cite{BRISQE_IQA} & 0.0126 & 0.0283 & 0.0100 & 0.0124 & 0.0094 & 0.0107 & 0.0009 & 0.0010 & 0.0196 & 0.0211 & 0.0370 & 0.1161 & 0.1345 & 0.1444 & 0.0149 & 0.0316 \\
 &  & RankIQA\cite{liu2017rankiqa} & 0.0109 & 0.0254 & 0.0091 & 0.0122 & 0.0095 & 0.0127 & 0.0009 & 0.0010 & 0.0202 & 0.0219 & 0.0411 & 0.1198 & 0.1361 & 0.1461 & 0.0153 & 0.0322 \\
 &  & DeepIQA\cite{DEEPIQ_IQA} & 0.0127 & 0.0284 & 0.0097 & 0.0117 & 0.0087 & 0.0117 & 0.0009 & 0.0010 & 0.0201 & 0.0210 & 0.0412 & 0.1198 & 0.1297 & 0.1416 & 0.0155 & 0.0323 \\
\cline{2-19}
 &  \multirow{14}{*}{\rotatebox[origin=c]{90}{FIQA}} & RankIQ\cite{RANKIQ_FIQA} & 0.0107 & 0.0247 & 0.0096 & 0.0118 & 0.0078 & 0.0107 & 0.0006 & 0.0008 & 0.0182 & 0.0200 & 0.0275 & 0.0402 & 0.1129 & 0.1292 & 0.0124 & 0.0181 \\
 &  & PFE\cite{PFE_FIQA} & 0.0090 & 0.0192 & 0.0089 & 0.0107 & 0.0090 & 0.0121 & 0.0009 & 0.0010 & 0.0185 & 0.0198 & 0.0316 & 0.0444 & \textbf{0.0941 (1)} & \textbf{0.1054 (1)} & 0.0130 & 0.0179 \\
 &  & SER-FIQ\cite{SERFIQ} & 0.0091 & 0.0207 & 0.0067 (3) & 0.0083 & 0.0035 (3) & \textbf{0.0053 (1)} & 0.0007 & 0.0008 & 0.0179 & 0.0192 & 0.0169 & 0.0308 & 0.1054* & 0.1217* (3) & 0.0091 & 0.0142 \\
 &  & FaceQnet\cite{hernandez2019faceqnet,faceqnetv1} & 0.0116 & 0.0254 & 0.0082 & 0.0101 & 0.0074 & 0.0099 & 0.0008 & 0.0009 & 0.0191 & 0.0204 & 0.0355 & 0.1066 & 0.1322 & 0.1459 & 0.0138 & 0.0289 \\
 &  & MagFace\cite{MagFace} & 0.0089 & 0.0198 & \textbf{0.0066 (1)} & 0.0082 (2) & 0.0048 & 0.0068 & \textbf{0.0006 (1)} & 0.0007 (3) & 0.0175 & 0.0185 (2) & 0.0219 & 0.0357 & 0.1257 & 0.1373 & 0.0101 & 0.0149 \\
 &  & SDD-FIQA\cite{SDDFIQA} & 0.0091 & 0.0212 & 0.0080 & 0.0098 & 0.0073 & 0.0097 & 0.0007 & 0.0008 & 0.0183 & 0.0197 & 0.0237 & 0.0413 & 0.1219 & 0.1372 & 0.0112 & 0.0171 \\
 &  & CR-FIQA(S) \cite{boutros_2023_crfiqa} & 0.0088 & 0.0211 & 0.0071 & 0.0089 & 0.0047 & 0.0071 & 0.0006 & 0.0008 & 0.0174 (2) & 0.0186 (3) & 0.0176 & 0.0317 & 0.1138 & 0.1301 & 0.0094 & 0.0147 \\
 &  & CR-FIQA(L) \cite{boutros_2023_crfiqa} & 0.0089 & 0.0189 & 0.0066 (2) & 0.0083 (3) & 0.0038 & 0.0056 & 0.0007 & 0.0009 & 0.0175 (3) & \textbf{0.0181 (1)} & 0.0161 & 0.0283 & 0.1043 (3) & 0.1279 & 0.0089 (3) & \textbf{0.0134 (1)} \\
 &  & DifFIQA(R) \cite{10449044} & 0.0087 & 0.0206 & 0.0087 & 0.0104 & 0.0034 (2) & 0.0054 (3) & 0.0006 (2) & 0.0007 (2) & 0.0178 & 0.0196 & \textbf{0.0158 (1)} & \textbf{0.0280 (1)} & 0.1109 & 0.1250 & 0.0092 & 0.0141 \\
 &  & eDifFIQA(L) \cite{babnikTBIOM2024} & \textbf{0.0079 (1)} & 0.0182 (2) & 0.0067 & \textbf{0.0081 (1)} & \textbf{0.0034 (1)} & 0.0053 (2) & 0.0006 (3) & \textbf{0.0007 (1)} & \textbf{0.0172 (1)} & 0.0186 & 0.0160 (3) & 0.0298 & 0.1129 & 0.1267 & \textbf{0.0086 (1)} & 0.0135 (2) \\
 &  & GraFIQs (S) \cite{grafiqs} & 0.0109 & 0.0227 & 0.0089 & 0.0122 & 0.0093 & 0.0122 & 0.0008 & 0.0009 & 0.0203 & 0.0219 & 0.0326 & 0.0459 & 0.1098 & 0.1313 & 0.0138 & 0.0193 \\
 &  & GraFIQs (L) \cite{grafiqs} & 0.0091 & \textbf{0.0181 (1)} & 0.0071 & 0.0096 & 0.0092 & 0.0122 & 0.0010 & 0.0011 & 0.0189 & 0.0204 & 0.0309 & 0.0437 & 0.0948 (2) & 0.1056 (2) & 0.0127 & 0.0175 \\
 &  & ViT-FIQA (T) (Ours) & 0.0079 (2) & 0.0188 (3) & 0.0071 & 0.0089 & 0.0037 & 0.0054 & 0.0006 & 0.0007 & 0.0181 & 0.0192 & 0.0160 (2) & 0.0281 (2) & 0.1096 & 0.1258 & 0.0089 (2) & 0.0135 (3) \\
 &  & ViT-FIQA (C) (Ours) & 0.0081 (3) & 0.0188 & 0.0075 & 0.0091 & 0.0038 & 0.0054 & 0.0007 & 0.0008 & 0.0186 & 0.0197 & 0.0161 & 0.0281 (3) & 0.1084 & 0.1248 & 0.0091 & 0.0136 \\
  \hline \hline 
\multirow{17}{*}{\rotatebox[origin=c]{90}{ElasticFace}} &  \multirow{3}{*}{\rotatebox[origin=c]{90}{IQA}} & BRISQUE\cite{BRISQE_IQA} & 0.0160 & 0.0302 & 0.0090 & 0.0100 & 0.0082 & 0.0107 & 0.0007 & 0.0010 & 0.0195 & 0.0203 & 0.0407 & 0.1043 & 0.1411 & 0.1637 & 0.0157 & 0.0294 \\
 &  & RankIQA\cite{liu2017rankiqa} & 0.0138 & 0.0274 & 0.0085 & 0.0096 & 0.0082 & 0.0105 & 0.0007 & 0.0010 & 0.0203 & 0.0209 & 0.0433 & 0.1086 & 0.1428 & 0.1661 & 0.0158 & 0.0297 \\
 &  & DeepIQA\cite{DEEPIQ_IQA} & 0.0162 & 0.0308 & 0.0088 & 0.0097 & 0.0074 & 0.0100 & 0.0008 & 0.0010 & 0.0201 & 0.0208 & 0.0431 & 0.1082 & 0.1379 & 0.1621 & 0.0161 & 0.0301 \\
\cline{2-19}
 &  \multirow{14}{*}{\rotatebox[origin=c]{90}{FIQA}} & RankIQ\cite{RANKIQ_FIQA} & 0.0139 & 0.0276 & 0.0089 & 0.0097 & 0.0067 & 0.0085 & 0.0005 (2) & 0.0008 & 0.0182 & 0.0188 & 0.0291 & 0.0394 & 0.1163 & 0.1342 & 0.0129 & 0.0175 \\
 &  & PFE\cite{PFE_FIQA} & 0.0109 & 0.0206 & 0.0077 & 0.0083 & 0.0078 & 0.0103 & 0.0008 & 0.0010 & 0.0184 & 0.0192 & 0.0340 & 0.0438 & 0.1012 (2) & \textbf{0.1107 (1)} & 0.0133 & 0.0172 \\
 &  & SER-FIQ\cite{SERFIQ} & 0.0114 & 0.0227 & 0.0064 & 0.0072 & 0.0031 & 0.0044 (3) & 0.0006 & 0.0008 & 0.0177 & 0.0184 & 0.0185 & 0.0292 & 0.1057* (3) & 0.1283* & 0.0096 & 0.0138 \\
 &  & FaceQnet\cite{hernandez2019faceqnet,faceqnetv1} & 0.0143 & 0.0274 & 0.0075 & 0.0082 & 0.0071 & 0.0084 & 0.0007 & 0.0009 & 0.0189 & 0.0196 & 0.0371 & 0.0951 & 0.1428 & 0.1645 & 0.0143 & 0.0266 \\
 &  & MagFace\cite{MagFace} & 0.0110 & 0.0211 & 0.0060 (2) & 0.0064 (2) & 0.0043 & 0.0059 & \textbf{0.0005 (1)} & 0.0007 (3) & 0.0173 & 0.0177 & 0.0237 & 0.0345 & 0.1331 & 0.1445 & 0.0105 & 0.0144 \\
 &  & SDD-FIQA\cite{SDDFIQA} & 0.0115 & 0.0231 & 0.0074 & 0.0080 & 0.0054 & 0.0067 & 0.0006 & 0.0008 & 0.0181 & 0.0186 & 0.0255 & 0.0377 & 0.1336 & 0.1564 & 0.0114 & 0.0158 \\
 &  & CR-FIQA(S) \cite{boutros_2023_crfiqa} & 0.0112 & 0.0223 & 0.0067 & 0.0073 & 0.0038 & 0.0057 & 0.0005 & 0.0008 & 0.0172 (3) & 0.0176 (3) & 0.0197 & 0.0301 & 0.1166 & 0.1411 & 0.0099 & 0.0140 \\
 &  & CR-FIQA(L) \cite{boutros_2023_crfiqa} & 0.0105 & 0.0206 & 0.0064 & 0.0069 & 0.0031 & 0.0045 & 0.0006 & 0.0009 & \textbf{0.0171 (1)} & \textbf{0.0175 (1)} & 0.0178 (3) & 0.0275 (3) & 0.1094 & 0.1265 (3) & 0.0092 (2) & 0.0130 (2) \\
 &  & DifFIQA(R) \cite{10449044} & 0.0113 & 0.0227 & 0.0076 & 0.0082 & 0.0031 & 0.0045 & 0.0006 & 0.0007 (2) & 0.0180 & 0.0186 & \textbf{0.0174 (1)} & \textbf{0.0271 (1)} & 0.1094 & 0.1279 & 0.0097 & 0.0136 \\
 &  & eDifFIQA(L) \cite{babnikTBIOM2024} & \textbf{0.0099 (1)} & 0.0205 & \textbf{0.0058 (1)} & \textbf{0.0062 (1)} & 0.0029 (2) & 0.0044 (2) & 0.0006 & \textbf{0.0007 (1)} & 0.0171 (2) & 0.0176 (2) & 0.0176 (2) & 0.0280 & 0.1129 & 0.1322 & \textbf{0.0090 (1)} & \textbf{0.0129 (1)} \\
 &  & GraFIQs (S) \cite{grafiqs} & 0.0132 & 0.0255 & 0.0086 & 0.0098 & 0.0079 & 0.0105 & 0.0007 & 0.0009 & 0.0202 & 0.0208 & 0.0351 & 0.0456 & 0.1142 & 0.1437 & 0.0143 & 0.0189 \\
 &  & GraFIQs (L) \cite{grafiqs} & 0.0107 & \textbf{0.0198 (1)} & 0.0066 & 0.0074 & 0.0080 & 0.0106 & 0.0009 & 0.0011 & 0.0189 & 0.0195 & 0.0329 & 0.0433 & \textbf{0.1005 (1)} & 0.1161 (2) & 0.0130 & 0.0169 \\
 &  & ViT-FIQA (T) (Ours) & 0.0101 (2) & 0.0203 (3) & 0.0064 (3) & 0.0068 (3) & 0.0030 (3) & 0.0045 & 0.0005 (3) & 0.0007 & 0.0175 & 0.0182 & 0.0180 & 0.0275 (2) & 0.1201 & 0.1503 & 0.0092 (3) & 0.0130 (3) \\
 &  & ViT-FIQA (C) (Ours) & 0.0101 (3) & 0.0202 (2) & 0.0067 & 0.0071 & \textbf{0.0028 (1)} & \textbf{0.0043 (1)} & 0.0006 & 0.0008 & 0.0181 & 0.0187 & 0.0181 & 0.0277 & 0.1169 & 0.1446 & 0.0094 & 0.0131 \\
  \hline \hline 
\multirow{17}{*}{\rotatebox[origin=c]{90}{MagFace}} &  \multirow{3}{*}{\rotatebox[origin=c]{90}{IQA}} & BRISQUE\cite{BRISQE_IQA} & 0.0147 & 0.0334 & 0.0101 & 0.0207 & 0.0117 & 0.0205 & 0.0009 & 0.0013 & 0.0199 & 0.0211 & 0.0676 & 0.1664 & 0.1601 & 0.1727 & 0.0208 & 0.0439 \\
 &  & RankIQA\cite{liu2017rankiqa} & 0.0128 & 0.0291 & 0.0092 & 0.0212 & 0.0119 & 0.0207 & 0.0009 & 0.0013 & 0.0208 & 0.0217 & 0.0518 & 0.1695 & 0.1619 & 0.1744 & 0.0179 & 0.0439 \\
 &  & DeepIQA\cite{DEEPIQ_IQA} & 0.0149 & 0.0335 & 0.0100 & 0.0204 & 0.0111 & 0.0199 & 0.0009 & 0.0013 & 0.0206 & 0.0215 & 0.0710 & 0.1691 & 0.1576 & 0.1706 & 0.0214 & 0.0443 \\
\cline{2-19}
 &  \multirow{14}{*}{\rotatebox[origin=c]{90}{FIQA}} & RankIQ\cite{RANKIQ_FIQA} & 0.0125 & 0.0302 & 0.0100 & 0.0199 & 0.0096 & 0.0178 & 0.0007 & 0.0010 & 0.0188 & 0.0198 & 0.0336 & 0.1133 & 0.1392 & 0.1514 & 0.0142 & 0.0337 \\
 &  & PFE\cite{PFE_FIQA} & 0.0103 & 0.0225 & 0.0087 & 0.0176 & 0.0115 & 0.0199 & 0.0009 & 0.0010 & 0.0190 & 0.0194 & 0.0380 & 0.0615 & 0.1199 (2) & \textbf{0.1333 (1)} & 0.0147 & 0.0237 \\
 &  & SER-FIQ\cite{SERFIQ} & 0.0107 & 0.0241 & 0.0074 & 0.0160 & 0.0045 (3) & 0.0099 & 0.0007 & 0.0011 & 0.0183 & 0.0187 & 0.0219 & 0.0541 & 0.1264* (3) & 0.1440* (3) & 0.0106 & 0.0207 \\
 &  & FaceQnet\cite{hernandez2019faceqnet,faceqnetv1} & 0.0133 & 0.0292 & 0.0082 & 0.0159 & 0.0096 & 0.0162 & 0.0008 & 0.0010 & 0.0193 & 0.0198 & 0.0602 & 0.1589 & 0.1584 & 0.1681 & 0.0186 & 0.0402 \\
 &  & MagFace\cite{MagFace} & 0.0100 & 0.0233 & \textbf{0.0066 (1)} & 0.0134 & 0.0057 & 0.0096 & 0.0006 (2) & \textbf{0.0008 (1)} & 0.0178 & 0.0184 & 0.0268 & 0.0579 & 0.1496 & 0.1603 & 0.0112 & 0.0205 \\
 &  & SDD-FIQA\cite{SDDFIQA} & 0.0106 & 0.0257 & 0.0081 & 0.0122 (2) & 0.0083 & 0.0128 & 0.0007 & 0.0008 & 0.0186 & 0.0194 & 0.0284 & 0.0834 & 0.1525 & 0.1656 & 0.0125 & 0.0257 \\
 &  & CR-FIQA(S) \cite{boutros_2023_crfiqa} & 0.0103 & 0.0246 & 0.0074 & 0.0137 & 0.0054 & 0.0068 (2) & 0.0007 & 0.0010 & 0.0177 (2) & 0.0182 (2) & 0.0225 & 0.0548 & 0.1339 & 0.1619 & 0.0106 & 0.0198 \\
 &  & CR-FIQA(L) \cite{boutros_2023_crfiqa} & 0.0100 & 0.0210 (2) & 0.0071 (3) & 0.0128 (3) & 0.0048 & \textbf{0.0061 (1)} & 0.0007 & 0.0008 & 0.0177 (3) & 0.0183 (3) & 0.0209 (3) & \textbf{0.0454 (1)} & 0.1296 & 0.1506 & 0.0102 (3) & \textbf{0.0174 (1)} \\
 &  & DifFIQA(R) \cite{10449044} & 0.0103 & 0.0251 & 0.0086 & 0.0171 & 0.0047 & 0.0104 & \textbf{0.0006 (1)} & 0.0008 & 0.0185 & 0.0194 & \textbf{0.0207 (1)} & 0.0598 & 0.1280 & 0.1501 & 0.0106 & 0.0221 \\
 &  & eDifFIQA(L) \cite{babnikTBIOM2024} & 0.0093 (3) & 0.0224 & 0.0067 (2) & \textbf{0.0113 (1)} & \textbf{0.0042 (1)} & 0.0097 & 0.0006 (3) & 0.0008 (2) & \textbf{0.0177 (1)} & \textbf{0.0182 (1)} & 0.0212 & 0.0596 & 0.1323 & 0.1498 & \textbf{0.0100 (1)} & 0.0203 \\
 &  & GraFIQs (S) \cite{grafiqs} & 0.0126 & 0.0268 & 0.0095 & 0.0211 & 0.0115 & 0.0202 & 0.0008 & 0.0011 & 0.0209 & 0.0219 & 0.0387 & 0.0637 & 0.1298 & 0.1556 & 0.0156 & 0.0258 \\
 &  & GraFIQs (L) \cite{grafiqs} & 0.0103 & \textbf{0.0203 (1)} & 0.0074 & 0.0174 & 0.0116 & 0.0181 & 0.0010 & 0.0014 & 0.0195 & 0.0202 & 0.0368 & 0.0617 & \textbf{0.1189 (1)} & 0.1374 (2) & 0.0144 & 0.0232 \\
 &  & ViT-FIQA (T) (Ours) & 0.0091 (2) & 0.0229 & 0.0073 & 0.0134 & 0.0046 & 0.0071 (3) & 0.0006 & 0.0008 (3) & 0.0182 & 0.0188 & 0.0208 (2) & 0.0454 (2) & 0.1316 & 0.1500 & 0.0101 (2) & 0.0181 (2) \\
 &  & ViT-FIQA (C) (Ours) & \textbf{0.0091 (1)} & 0.0221 (3) & 0.0077 & 0.0163 & 0.0045 (2) & 0.0090 & 0.0007 & 0.0008 & 0.0188 & 0.0194 & 0.0211 & 0.0459 (3) & 0.1286 & 0.1497 & 0.0103 & 0.0189 (3) \\
  \hline \hline 
\multirow{17}{*}{\rotatebox[origin=c]{90}{SwinFace}} &  \multirow{3}{*}{\rotatebox[origin=c]{90}{IQA}} & BRISQUE\cite{BRISQE_IQA} & 0.0204 & 0.0470 & 0.0124 & 0.0150 & 0.0199 & 0.0692 & 0.0010 & 0.0014 & 0.0197 & 0.0261 & 0.0526 & 0.0846 & 0.1373 & 0.1556 & 0.0210 & 0.0405 \\
 &  & RankIQA\cite{liu2017rankiqa} & 0.0181 & 0.0431 & 0.0120 & 0.0146 & 0.0225 & 0.0705 & 0.0010 & 0.0014 & 0.0202 & 0.0266 & 0.0551 & 0.0890 & 0.1388 & 0.1571 & 0.0215 & 0.0409 \\
 &  & DeepIQA\cite{DEEPIQ_IQA} & 0.0204 & 0.0471 & 0.0115 & 0.0143 & 0.0200 & 0.0642 & 0.0011 & 0.0013 & 0.0203 & 0.0263 & 0.0560 & 0.0885 & 0.1346 & 0.1531 & 0.0215 & 0.0403 \\
\cline{2-19}
 &  \multirow{14}{*}{\rotatebox[origin=c]{90}{FIQA}} & RankIQ\cite{RANKIQ_FIQA} & 0.0182 & 0.0460 & 0.0115 & 0.0141 & 0.0201 & 0.0658 & 0.0007 & 0.0010 & 0.0185 & 0.0240 & 0.0412 & 0.0557 & 0.1072 & 0.1362 & 0.0184 & 0.0344 \\
 &  & PFE\cite{PFE_FIQA} & 0.0141 & 0.0335 & 0.0101 & 0.0118 & 0.0196 & 0.0645 & 0.0010 & 0.0012 & 0.0188 & 0.0237 & 0.0454 & 0.0600 & 0.0933 (2) & 0.1115 (2) & 0.0182 & 0.0324 \\
 &  & SER-FIQ\cite{SERFIQ} & 0.0147 & 0.0351 & 0.0083 & 0.0108 & 0.0131 & 0.0567 & 0.0008 & 0.0011 & 0.0180 & \textbf{0.0198 (1)} & 0.0306 & 0.0458 & 0.0993* & 0.1252* & 0.0142 & 0.0282 \\
 &  & FaceQnet\cite{hernandez2019faceqnet,faceqnetv1} & 0.0185 & 0.0447 & 0.0089 & 0.0119 & 0.0171 & 0.0608 & 0.0009 & 0.0010 & 0.0193 & 0.0246 & 0.0486 & 0.0777 & 0.1344 & 0.1576 & 0.0189 & 0.0368 \\
 &  & MagFace\cite{MagFace} & 0.0142 & 0.0353 & 0.0078 (2) & 0.0098 (2) & 0.0132 & 0.0559 & \textbf{0.0006 (1)} & \textbf{0.0007 (1)} & \textbf{0.0171 (1)} & 0.0216 & 0.0348 & 0.0502 & 0.1233 & 0.1497 & 0.0146 & 0.0289 \\
 &  & SDD-FIQA\cite{SDDFIQA} & 0.0158 & 0.0399 & 0.0083 & 0.0107 & 0.0161 & 0.0602 & 0.0007 & 0.0009 & 0.0182 & 0.0234 & 0.0373 & 0.0539 & 0.1302 & 0.1488 & 0.0161 & 0.0315 \\
 &  & CR-FIQA(S) \cite{boutros_2023_crfiqa} & 0.0153 & 0.0386 & 0.0087 & 0.0108 & 0.0120 (2) & \textbf{0.0383 (1)} & 0.0007 & 0.0010 & 0.0172 & 0.0223 & 0.0317 & 0.0475 & 0.1082 & 0.1384 & 0.0142 & \textbf{0.0264 (1)} \\
 &  & CR-FIQA(L) \cite{boutros_2023_crfiqa} & 0.0133 (2) & 0.0312 (2) & 0.0079 (3) & 0.0103 (3) & 0.0124 & 0.0558 & 0.0007 & 0.0008 & 0.0172 (3) & 0.0215 & 0.0298 (3) & 0.0451 (2) & 0.1052 & 0.1348 & 0.0136 (2) & 0.0275 (2) \\
 &  & DifFIQA(R) \cite{10449044} & 0.0154 & 0.0380 & 0.0098 & 0.0116 & 0.0122 & 0.0552 & 0.0006 (2) & 0.0008 & 0.0180 & 0.0234 & \textbf{0.0290 (1)} & \textbf{0.0443 (1)} & 0.1035 & 0.1271 & 0.0142 & 0.0289 \\
 &  & eDifFIQA(L) \cite{babnikTBIOM2024} & 0.0139 & 0.0337 & \textbf{0.0077 (1)} & \textbf{0.0097 (1)} & \textbf{0.0118 (1)} & 0.0544 (2) & 0.0006 (3) & 0.0008 (2) & 0.0171 (2) & 0.0222 & 0.0297 (2) & 0.0454 (3) & 0.1000 & 0.1339 & \textbf{0.0135 (1)} & 0.0277 (3) \\
 &  & GraFIQs (S) \cite{grafiqs} & 0.0163 & 0.0393 & 0.0122 & 0.0148 & 0.0194 & 0.0678 & 0.0009 & 0.0012 & 0.0205 & 0.0269 & 0.0471 & 0.0630 & 0.1040 & 0.1341 & 0.0194 & 0.0355 \\
 &  & GraFIQs (L) \cite{grafiqs} & 0.0134 (3) & \textbf{0.0309 (1)} & 0.0083 & 0.0119 & 0.0199 & 0.0670 & 0.0010 & 0.0014 & 0.0191 & 0.0248 & 0.0447 & 0.0596 & 0.0941 (3) & \textbf{0.1084 (1)} & 0.0177 & 0.0326 \\
 &  & ViT-FIQA (T) (Ours) & 0.0139 & 0.0363 & 0.0086 & 0.0109 & 0.0120 (3) & 0.0549 (3) & 0.0007 & 0.0008 (3) & 0.0177 & 0.0202 (2) & 0.0302 & 0.0458 & 0.1015 & 0.1292 & 0.0138 (3) & 0.0281 \\
 &  & ViT-FIQA (C) (Ours) & \textbf{0.0132 (1)} & 0.0333 (3) & 0.0091 & 0.0111 & 0.0121 & 0.0556 & 0.0007 & 0.0008 & 0.0180 & 0.0214 (3) & 0.0301 & 0.0460 & \textbf{0.0924 (1)} & 0.1164 (3) & 0.0139 & 0.0280 \\
  \hline \hline 
\multirow{17}{*}{\rotatebox[origin=c]{90}{TransFace}} &  \multirow{3}{*}{\rotatebox[origin=c]{90}{IQA}} & BRISQUE\cite{BRISQE_IQA} & 0.0122 & 0.0248 & 0.0085 & 0.0101 & 0.0064 & 0.0082 & 0.0007 & 0.0008 & 0.0194 & 0.2144 & 0.0314 & 0.1213 & 0.0663 & 0.0811 & 0.0131 & 0.0633 \\
 &  & RankIQA\cite{liu2017rankiqa} & 0.0109 & 0.0224 & 0.0083 & 0.0100 & 0.0063 & 0.0079 & 0.0007 & 0.0008 & 0.0203 & 0.2159 & 0.0328 & 0.1250 & 0.0667 & 0.0818 & 0.0132 & 0.0636 \\
 &  & DeepIQA\cite{DEEPIQ_IQA} & 0.0122 & 0.0249 & 0.0085 & 0.0102 & 0.0056 & 0.0076 & 0.0007 & 0.0008 & 0.0200 & 0.1181 (2) & 0.0326 & 0.1242 & 0.0648 & 0.0796 & 0.0133 & 0.0476 \\
\cline{2-19}
 &  \multirow{14}{*}{\rotatebox[origin=c]{90}{FIQA}} & RankIQ\cite{RANKIQ_FIQA} & 0.0105 & 0.0222 & 0.0084 & 0.0098 & 0.0053 & 0.0069 & 0.0006 (3) & 0.0007 (3) & 0.0183 & 0.2130 & 0.0247 & 0.0357 & 0.0482 & 0.0676 & 0.0113 & 0.0481 \\
 &  & PFE\cite{PFE_FIQA} & 0.0087 & 0.0169 (3) & 0.0077 & 0.0089 & 0.0063 & 0.0078 & 0.0007 & 0.0008 & 0.0186 & 0.2119 & 0.0272 & 0.0376 & 0.0463 & 0.0635 & 0.0115 & 0.0473 \\
 &  & SER-FIQ\cite{SERFIQ} & 0.0091 & 0.0181 & 0.0056 (2) & 0.0073 & 0.0026 & 0.0038 & 0.0007 & 0.0008 & 0.0178 & 0.2131 & 0.0151 & 0.0258 & 0.0400* (3) & 0.0539* & 0.0085 & 0.0448 \\
 &  & FaceQnet\cite{hernandez2019faceqnet,faceqnetv1} & 0.0111 & 0.0218 & 0.0072 & 0.0086 & 0.0053 & 0.0068 & 0.0006 & 0.0007 & 0.0191 & 0.2121 & 0.0287 & 0.1142 & 0.0597 & 0.0867 & 0.0120 & 0.0607 \\
 &  & MagFace\cite{MagFace} & 0.0089 & 0.0183 & 0.0061 (3) & 0.0069 (2) & 0.0034 & 0.0049 & \textbf{0.0006 (1)} & \textbf{0.0007 (1)} & 0.0176 & 0.2117 & 0.0187 & 0.0291 & 0.0529 & 0.0656 & 0.0092 & 0.0453 \\
 &  & SDD-FIQA\cite{SDDFIQA} & 0.0094 & 0.0195 & 0.0071 & 0.0082 & 0.0048 & 0.0064 & 0.0006 & 0.0007 & 0.0183 & 0.2122 & 0.0191 & 0.0303 & 0.0603 & 0.0767 & 0.0099 & 0.0462 \\
 &  & CR-FIQA(S) \cite{boutros_2023_crfiqa} & 0.0092 & 0.0188 & 0.0061 & 0.0072 & 0.0031 & 0.0044 & 0.0006 & 0.0007 & \textbf{0.0172 (1)} & 0.2113 & 0.0150 & 0.0259 & 0.0474 & 0.0663 & 0.0085 & 0.0447 \\
 &  & CR-FIQA(L) \cite{boutros_2023_crfiqa} & 0.0087 & 0.0170 & 0.0061 & 0.0072 (3) & 0.0027 & 0.0037 (3) & 0.0007 & 0.0008 & 0.0176 (3) & \textbf{0.0766 (1)} & 0.0144 & 0.0253 & 0.0510 & 0.0662 & 0.0084 (3) & \textbf{0.0217 (1)} \\
 &  & DifFIQA(R) \cite{10449044} & 0.0093 & 0.0193 & 0.0073 & 0.0082 & 0.0024 & 0.0037 & 0.0006 (2) & 0.0007 (2) & 0.0179 & 0.2124 & 0.0143 (3) & 0.0249 (3) & 0.0410 & 0.0544 & 0.0086 & 0.0449 \\
 &  & eDifFIQA(L) \cite{babnikTBIOM2024} & 0.0087 (3) & 0.0177 & \textbf{0.0055 (1)} & \textbf{0.0063 (1)} & \textbf{0.0023 (1)} & 0.0036 (2) & 0.0006 & 0.0007 & 0.0172 (2) & 0.2112 (3) & 0.0140 (2) & 0.0248 (2) & 0.0411 & 0.0531 (3) & \textbf{0.0080 (1)} & 0.0440 (2) \\
 &  & GraFIQs (S) \cite{grafiqs} & 0.0105 & 0.0205 & 0.0082 & 0.0099 & 0.0061 & 0.0079 & 0.0008 & 0.0009 & 0.0202 & 0.2162 & 0.0288 & 0.0397 & 0.0504 & 0.0651 & 0.0124 & 0.0492 \\
 &  & GraFIQs (L) \cite{grafiqs} & 0.0090 & \textbf{0.0166 (1)} & 0.0064 & 0.0079 & 0.0060 & 0.0077 & 0.0008 & 0.0009 & 0.0190 & 0.2145 & 0.0263 & 0.0373 & 0.0473 & 0.0613 & 0.0113 & 0.0475 \\
 &  & ViT-FIQA (T) (Ours) & 0.0084 (2) & 0.0174 & 0.0065 & 0.0074 & 0.0024 (3) & 0.0037 & 0.0006 & 0.0007 & 0.0180 & 0.2113 & \textbf{0.0139 (1)} & \textbf{0.0242 (1)} & 0.0373 (2) & 0.0500 (2) & 0.0083 (2) & 0.0441 (3) \\
 &  & ViT-FIQA (C) (Ours) & \textbf{0.0083 (1)} & 0.0167 (2) & 0.0067 & 0.0076 & 0.0024 (2) & \textbf{0.0035 (1)} & 0.0006 & 0.0007 & 0.0185 & 0.2116 & 0.0143 & 0.0250 & \textbf{0.0364 (1)} & \textbf{0.0490 (1)} & 0.0085 & 0.0442 \\
  \hline \hline 
\end{tabular}}
\end{center}
\vspace{-6mm}
\caption{The pAUCs (30\%) of EDC achieved by our ViT-FIQA and the SOTA methods under different FR experimental settings. The notions of $1e-3$ and $1e-4$ indicate the value of the fixed FMR at which the EDC curves (FNMR vs.~reject) were calculated. The results are compared to three IQA and twelve FIQA approaches. The XQLFW dataset uses SER-FIQ (marked with *) as FIQ labeling method.}
\vspace{3mm}
 \label{tab:pauc_all}
 \vspace{-6mm}
\end{table*}

\paragraph{Comparison on CNN-based FR:}
Tables~\ref{tab:auc_all} and ~\ref{tab:pauc_all} (first four blocks) present a comprehensive comparison of our proposed ViT-FIQA methods against a diverse set of SOTA FIQA and IQA methods, evaluated under four reference FR backbones (ArcFace, CurricularFace, ElasticFace, and MagFace) across seven benchmarks. The AUC and pAUC (30\%) values of the Error vs. Reject Curves (ERC) are reported at two FMR thresholds: $1e-3$ and $1e-4$. 

Our results highlight several important patterns. First, IQA baselines consistently underperform compared to all FIQA approaches, confirming that general-purpose perceptual quality does not capture the discriminative utility of facial images in recognition pipelines. This trend is uniform across all FR backbones and operating points.

Second, our proposed ViT-FIQA (T) and ViT-FIQA (C) show strong and stable behavior across the board. They achieve multiple top-3 positions in the rankings, particularly on benchmarks such as Adience, CFP-FP, CPLFW, and XQLFW. This demonstrates that the transformer-based design is competitive with the most recent SOTA approaches, and often closes the gap even under stricter operating points ($1e-4$). Importantly, the methods show consistent transferability across different FR backbones, underlining their robustness as FIQ estimators.  

Third, a recurring pattern is that our methods tend to perform best on CFP-FP, CPLFW and Adience, where ViT-FIQA (T/C) often rank first or second. In contrast, our methods most often fall short on AgeDB-30 and CALFW, with CR-FIQA(L) and eDifFIQA (L) regularly achieving the best results. This suggests that while our models effectively capture intra-class variability caused by pose and image quality, competing methods may be better suited to handle the subtler degradations introduced by large age gaps on specific benchmarks.

Finally, although CR-FIQA(L) and eDifFIQA(L) often yield the lowest mean pAUC values, our methods consistently remain competitive and within a narrow margin. In several cases, ViT-FIQA (T/C) even tie with or surpass these approaches, particularly on challenging subsets and when evaluated under stricter operating points, highlighting the potential on challenging and maintaining strong overall competitiveness across backbones.

\begin{figure*}[bh!]
\centering
\includegraphics[width=0.95\textwidth, trim=1 1 1 1,clip]{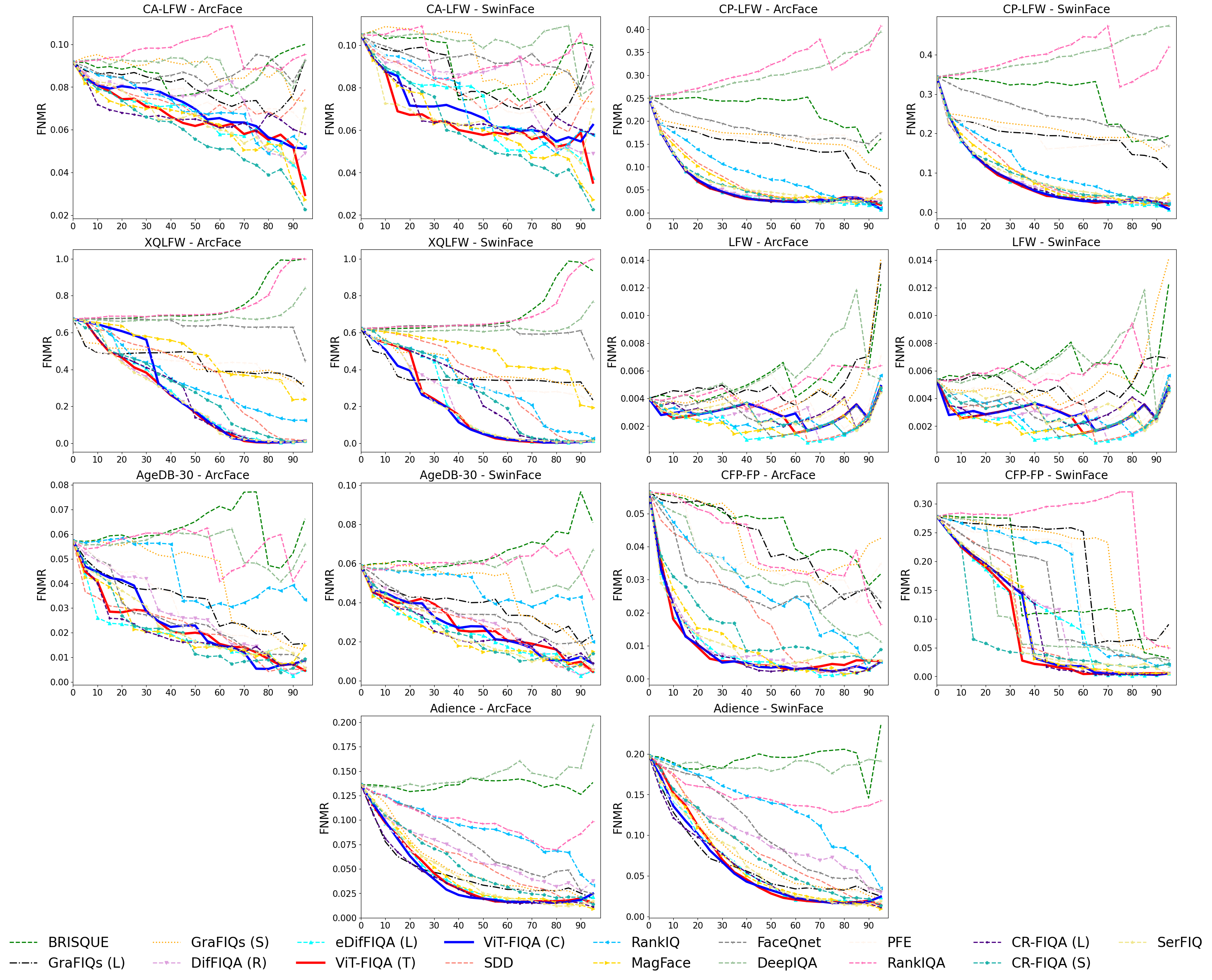}
\vspace{-2mm}
\caption{EDC curves for FNMR@FMR=1e-4 for all evaluated benchmarks using CNN-based (ArcFace) and ViT-based (SwinFace) FR models. The proposed ViT-FIQA (T) is shown in \textcolor{red}{solid red}, ViT-FIQA (C) is shown in \textcolor{blue}{solid blue}.}
\label{fig:fiqa_cnn_fmr4}
\vspace{-3mm}
\end{figure*}

\paragraph{Comparison on ViT-based FR:}
Tables~\ref{tab:auc_all} and ~\ref{tab:pauc_all} (last two blocks) present the AUCs and pAUCs (30\%) of EDC under ViT-based FR backbones. As in the CNN-based experiments, IQA baselines perform considerably worse than all FIQA approaches, confirming once again that generic image quality measures are misaligned with FR performance.  

For SwinFace, our proposed methods perform strongly. ViT-FIQA (C) achieves the best performance on Adience at both operating points (exception made for pAUC under FMR=$1e-4$), and provides the lowest values on XQLFW with the same pattern, outperforming all competing methods including SER-FIQ. ViT-FIQA (T) follows closely, ranking second or third on Adience, CFP-FP, and XQLFW. On CFP-FP at $1e-3$, both variants are competitive but slightly behind CR-FIQA and eDifFIQA. Overall, the mean pAUC values of ViT-FIQA (T) are always in the top 3 positions under this backbone.  

For TransFace, ViT-FIQA (C) achieves the best performance on Adience at both operating points (exception made for pAUC under FMR=$1e-4$) and ranks first on XQLFW, while ViT-FIQA (T) secures second or third place on the same benchmarks. On CFP-FP, both variants remain extremely competitive with CR-FIQA(L) and eDifFIQA (L). In terms of mean pAUC, our ViT-FIQA (T) consistently achieves the second or third best results, tightly following CR-FIQA (L)  and eDifFIQA (L).

A consistent pattern emerges across both ViT-based backbones: our methods excel on Adience and XQLFW, often taking the top rank, and they maintain strong competitiveness on pose-related benchmarks such as CFP-FP and CPLFW. In contrast, on AgeDB-30 and CALFW, ViT-FIQA (T/C) regularly fall behind CR-FIQA (L) and eDifFIQA (L), echoing the trend observed in the CNN-based experiments and confirming that our approach generalizes effectively across both CNN- and ViT-based recognition models, while also indicating that age-related variability remains the most challenging condition.

\vspace{-2mm}
\section{Conclusions}
\vspace{-1mm}

In this work, we introduced ViT-FIQA, a novel Vision Transformer-based approach for FIQA. By augmenting standard ViT architectures with a dedicated, learnable quality token, our method effectively captures global contextual information across facial regions and predicts a scalar quality score aligned with the FR task. The integration of a classifiability-related loss ensures that the learned quality scores reflect the relative discriminability of facial images, maintaining a tight coupling with the underlying recognition pipeline.
Our extensive experiments across multiple benchmarks and diverse FR models, spanning both CNN-based and transformer-based architectures, demonstrate the robustness and generalizability of ViT-FIQA. The method consistently ranks among the top performers, particularly excelling in challenging datasets like Adience, CFP-FP, CPLFW and XQLFW. These results highlight the potential of leveraging transformer-based architectures for FIQA, while highlighting limitations on specific benchmarks like AgeDB-30 and CALFW.

\paragraph{Acknowledgments}
This research work has been funded by the German Federal Ministry of Education and Research and the Hessen State Ministry for Higher Education, Research and the Arts within their joint support of the National Research Center for Applied Cybersecurity ATHENE. This work was carried out during the tenure of an ERCIM ’Alain Bensoussan‘ Fellowship Programme.

{
    \small
    \bibliographystyle{ieeenat_fullname}
    \bibliography{main}

\begin{thebibliography}{54}
\providecommand{\natexlab}[1]{#1}
\providecommand{\url}[1]{\texttt{#1}}
\expandafter\ifx\csname urlstyle\endcsname\relax
  \providecommand{\doi}[1]{doi: #1}\else
  \providecommand{\doi}{doi: \begingroup \urlstyle{rm}\Url}\fi

\bibitem[Babnik et~al.(2023{\natexlab{a}})Babnik, Damer, and {\v{S}}truc]{DBLP:conf/iwbf/BabnikDS23}
{\v{Z}}iga Babnik, Naser Damer, and Vitomir {\v{S}}truc.
\newblock Optimization-based improvement of face image quality assessment techniques.
\newblock In \emph{11th International Workshop on Biometrics and Forensics, {IWBF} 2023, Barcelona, Spain, April 19-20, 2023}, pages 1--6. {IEEE}, 2023{\natexlab{a}}.

\bibitem[Babnik et~al.(2023{\natexlab{b}})Babnik, Peer, and {\v{S}}truc]{10449044}
{\v{Z}}iga Babnik, Peter Peer, and Vitomir {\v{S}}truc.
\newblock Diffiqa: Face image quality assessment using denoising diffusion probabilistic models.
\newblock In \emph{2023 IEEE International Joint Conference on Biometrics (IJCB)}, pages 1--10, 2023{\natexlab{b}}.

\bibitem[Babnik et~al.(2024)Babnik, Peer, and {\v{S}}truc]{babnikTBIOM2024}
{\v{Z}}iga Babnik, Peter Peer, and Vitomir {\v{S}}truc.
\newblock {eDifFIQA: Towards Efficient Face Image Quality Assessment based on Denoising Diffusion Probabilistic Models}.
\newblock \emph{IEEE Transactions on Biometrics, Behavior, and Identity Science (TBIOM)}, 2024.

\bibitem[Best{-}Rowden and Jain(2018)]{best2018learning}
Lacey Best{-}Rowden and Anil~K. Jain.
\newblock Learning face image quality from human assessments.
\newblock \emph{{IEEE} Trans. Inf. Forensics Secur.}, 13\penalty0 (12):\penalty0 3064--3077, 2018.

\bibitem[Bosse et~al.(2018)Bosse, Maniry, M{\"{u}}ller, Wiegand, and Samek]{DEEPIQ_IQA}
Sebastian Bosse, Dominique Maniry, Klaus{-}Robert M{\"{u}}ller, Thomas Wiegand, and Wojciech Samek.
\newblock Deep neural networks for no-reference and full-reference image quality assessment.
\newblock \emph{{IEEE} Trans. Image Process.}, 27\penalty0 (1):\penalty0 206--219, 2018.

\bibitem[Boutros et~al.(2022)Boutros, Damer, Kirchbuchner, and Kuijper]{elasticface}
Fadi Boutros, Naser Damer, Florian Kirchbuchner, and Arjan Kuijper.
\newblock Elasticface: Elastic margin loss for deep face recognition.
\newblock In \emph{{IEEE/CVF} Conference on Computer Vision and Pattern Recognition Workshops, {CVPR} Workshops 2022, New Orleans, LA, USA, June 19-20, 2022}, pages 1577--1586. {IEEE}, 2022.

\bibitem[Boutros et~al.(2023)Boutros, Fang, Klemt, Fu, and Damer]{boutros_2023_crfiqa}
Fadi Boutros, Meiling Fang, Marcel Klemt, Biying Fu, and Naser Damer.
\newblock {CR-FIQA:} face image quality assessment by learning sample relative classifiability.
\newblock In \emph{{IEEE/CVF} Conference on Computer Vision and Pattern Recognition, {CVPR} 2023, Vancouver, BC, Canada, June 17-24, 2023}, pages 5836--5845. {IEEE}, 2023.

\bibitem[Chen et~al.(2015)Chen, Deng, Bai, and Su]{RANKIQ_FIQA}
Jiansheng Chen, Yu Deng, Gaocheng Bai, and Guangda Su.
\newblock Face image quality assessment based on learning to rank.
\newblock \emph{{IEEE} Signal Process. Lett.}, 22\penalty0 (1):\penalty0 90--94, 2015.

\bibitem[Dan et~al.(2023)Dan, Liu, Xie, Deng, Xie, Xie, and Sun]{Dan_2023_TransFace}
Jun Dan, Yang Liu, Haoyu Xie, Jiankang Deng, Haoran Xie, Xuansong Xie, and Baigui Sun.
\newblock Transface: Calibrating transformer training for face recognition from a data-centric perspective.
\newblock In \emph{ICCV}, pages 20585--20596, 2023.

\bibitem[Deng et~al.(2019)Deng, Guo, Xue, and Zafeiriou]{deng2019arcface}
Jiankang Deng, Jia Guo, Niannan Xue, and Stefanos Zafeiriou.
\newblock Arcface: Additive angular margin loss for deep face recognition.
\newblock In \emph{{IEEE} Conference on Computer Vision and Pattern Recognition, {CVPR} 2019, Long Beach, CA, USA, June 16-20, 2019}, pages 4690--4699. Computer Vision Foundation / {IEEE}, 2019.

\bibitem[Dosovitskiy et~al.(2021)Dosovitskiy, Beyer, Kolesnikov, Weissenborn, Zhai, Unterthiner, Dehghani, Minderer, Heigold, Gelly, Uszkoreit, and Houlsby]{DBLP:conf/iclr/DosovitskiyB0WZ21}
Alexey Dosovitskiy, Lucas Beyer, Alexander Kolesnikov, Dirk Weissenborn, Xiaohua Zhai, Thomas Unterthiner, Mostafa Dehghani, Matthias Minderer, Georg Heigold, Sylvain Gelly, Jakob Uszkoreit, and Neil Houlsby.
\newblock An image is worth 16x16 words: Transformers for image recognition at scale.
\newblock In \emph{9th International Conference on Learning Representations, {ICLR} 2021, Virtual Event, Austria, May 3-7, 2021}. OpenReview.net, 2021.

\bibitem[Eidinger et~al.(2014)Eidinger, Enbar, and Hassner]{Adience}
Eran Eidinger, Roee Enbar, and Tal Hassner.
\newblock Age and gender estimation of unfiltered faces.
\newblock \emph{{IEEE} Trans. Inf. Forensics Secur.}, 9\penalty0 (12):\penalty0 2170--2179, 2014.

\bibitem[Frontex(2015)]{frontex2015best}
Frontex.
\newblock Best practice technical guidelines for automated border control (abc) systems, 2015.

\bibitem[Fu et~al.(2022)Fu, Chen, Henniger, and Damer]{BiyingWACV}
Biying Fu, Cong Chen, Olaf Henniger, and Naser Damer.
\newblock A deep insight into measuring face image utility with general and face-specific image quality metrics.
\newblock In \emph{{IEEE/CVF} Winter Conference on Applications of Computer Vision, {WACV} 2022, Waikoloa, HI, USA, January 3-8, 2022}, pages 1121--1130. {IEEE}, 2022.

\bibitem[Grother and Tabassi(2007)]{GT07}
P. Grother and E. Tabassi.
\newblock Performance of biometric quality measures.
\newblock \emph{IEEE Trans.~on Pattern Analysis and Machine Intelligence}, 29\penalty0 (4):\penalty0 531--543, 2007.

\bibitem[Grother et~al.(Sep. 2021)Grother, A.~Hom, and Hanaoka]{NISTQuaity}
P. Grother, M.~Ngan A.~Hom, and K. Hanaoka.
\newblock Ongoing face recognition vendor test (frvt) part 5: Face image quality assessment (4th draft).
\newblock In \emph{National Institute of Standards and Technology}. Tech. Rep., Sep. 2021.

\bibitem[Guo et~al.(2016)Guo, Zhang, Hu, He, and Gao]{guo_2016_ms1m}
Yandong Guo, Lei Zhang, Yuxiao Hu, Xiaodong He, and Jianfeng Gao.
\newblock Ms-celeb-1m: {A} dataset and benchmark for large-scale face recognition.
\newblock In \emph{Computer Vision - {ECCV} 2016 - 14th European Conference, Amsterdam, The Netherlands, October 11-14, 2016, Proceedings, Part {III}}, pages 87--102. Springer, 2016.

\bibitem[Hernandez{-}Ortega et~al.(2019)Hernandez{-}Ortega, Galbally, Fi{\'{e}}rrez, Haraksim, and Beslay]{hernandez2019faceqnet}
Javier Hernandez{-}Ortega, Javier Galbally, Julian Fi{\'{e}}rrez, Rudolf Haraksim, and Laurent Beslay.
\newblock Faceqnet: Quality assessment for face recognition based on deep learning.
\newblock In \emph{2019 International Conference on Biometrics, {ICB} 2019, Crete, Greece, June 4-7, 2019}, pages 1--8. {IEEE}, 2019.

\bibitem[Hernandez{-}Ortega et~al.(2020)Hernandez{-}Ortega, Galbally, Fi{\'{e}}rrez, and Beslay]{faceqnetv1}
Javier Hernandez{-}Ortega, Javier Galbally, Julian Fi{\'{e}}rrez, and Laurent Beslay.
\newblock Biometric quality: Review and application to face recognition with faceqnet.
\newblock \emph{CoRR}, abs/2006.03298, 2020.

\bibitem[Huang et~al.(2007)Huang, Ramesh, Berg, and Learned-Miller]{LFWTech}
Gary~B. Huang, Manu Ramesh, Tamara Berg, and Erik Learned-Miller.
\newblock Labeled faces in the wild: A database for studying face recognition in unconstrained environments.
\newblock Technical Report 07-49, University of Massachusetts, Amherst, 2007.

\bibitem[Huang et~al.(2020)Huang, Wang, Tai, Liu, Shen, Li, Li, and Huang]{curricularFace}
Yuge Huang, Yuhan Wang, Ying Tai, Xiaoming Liu, Pengcheng Shen, Shaoxin Li, Jilin Li, and Feiyue Huang.
\newblock Curricularface: Adaptive curriculum learning loss for deep face recognition.
\newblock In \emph{2020 {IEEE/CVF} Conference on Computer Vision and Pattern Recognition, {CVPR} 2020, Seattle, WA, USA, June 13-19, 2020}, pages 5900--5909. Computer Vision Foundation / {IEEE}, 2020.

\bibitem[{ISO/IEC JTC1 SC37 Biometrics}(2016)]{ISOIEC29794-1}
{ISO/IEC JTC1 SC37 Biometrics}.
\newblock {ISO/IEC 29794-1:2016 Information technology - Biometric sample quality - Part 1: Framework}.
\newblock International Organization for Standardization, 2016.

\bibitem[{ISO/IEC JTC1 SC37 Biometrics}(2021)]{iso_metric}
{ISO/IEC JTC1 SC37 Biometrics}.
\newblock {ISO/IEC 19795-1:2021 Information technology — Biometric performance testing and reporting — Part 1: Principles and framework}.
\newblock International Organization for Standardization, 2021.

\bibitem[Kim et~al.(2023)Kim, Park, Kang, and Woo]{Kim_2023_SViT}
Geunsu Kim, Gyudo Park, Soohyeok Kang, and Simon~S. Woo.
\newblock S-vit: Sparse vision transformer for accurate face recognition.
\newblock In \emph{ACM SAC}, pages 1130--1138, 2023.

\bibitem[Kim et~al.(2024)Kim, Su, Liu, Jain, and Liu]{DBLP:conf/cvpr/KimS0JL24}
Minchul Kim, Yiyang Su, Feng Liu, Anil Jain, and Xiaoming Liu.
\newblock Keypoint relative position encoding for face recognition.
\newblock In \emph{{IEEE/CVF} Conference on Computer Vision and Pattern Recognition, {CVPR} 2024, Seattle, WA, USA, June 16-22, 2024}, pages 244--255. {IEEE}, 2024.

\bibitem[Kirillov et~al.(2023)Kirillov, Mintun, Ravi, Mao, Rolland, Gustafson, Xiao, Whitehead, Berg, Lo, Doll{\'{a}}r, and Girshick]{DBLP:conf/iccv/KirillovMRMRGXW23}
Alexander Kirillov, Eric Mintun, Nikhila Ravi, Hanzi Mao, Chlo{\'{e}} Rolland, Laura Gustafson, Tete Xiao, Spencer Whitehead, Alexander~C. Berg, Wan{-}Yen Lo, Piotr Doll{\'{a}}r, and Ross~B. Girshick.
\newblock Segment anything.
\newblock In \emph{{IEEE/CVF} International Conference on Computer Vision, {ICCV} 2023, Paris, France, October 1-6, 2023}, pages 3992--4003. {IEEE}, 2023.

\bibitem[Knoche et~al.(2021)Knoche, H{\"{o}}rmann, and Rigoll]{XQLFW}
Martin Knoche, Stefan H{\"{o}}rmann, and Gerhard Rigoll.
\newblock Cross-quality {LFW:} {A} database for analyzing cross- resolution image face recognition in unconstrained environments.
\newblock In \emph{16th {IEEE} International Conference on Automatic Face and Gesture Recognition, {FG} 2021, Jodhpur, India, December 15-18, 2021}, pages 1--5. {IEEE}, 2021.

\bibitem[Kolf et~al.(2024)Kolf, Damer, and Boutros]{grafiqs}
Jan~Niklas Kolf, Naser Damer, and Fadi Boutros.
\newblock Grafiqs: Face image quality assessment using gradient magnitudes.
\newblock In \emph{2024 IEEE/CVF Conference on Computer Vision and Pattern Recognition Workshops (CVPRW)}, pages 1490--1499, 2024.

\bibitem[Liu et~al.(2017)Liu, van~de Weijer, and Bagdanov]{liu2017rankiqa}
Xialei Liu, Joost van~de Weijer, and Andrew~D. Bagdanov.
\newblock Rankiqa: Learning from rankings for no-reference image quality assessment.
\newblock In \emph{{IEEE} International Conference on Computer Vision, {ICCV} 2017, Venice, Italy, October 22-29, 2017}, pages 1040--1049. {IEEE} Computer Society, 2017.

\bibitem[Loshchilov and Hutter(2019)]{DBLP:conf/iclr/LoshchilovH19}
Ilya Loshchilov and Frank Hutter.
\newblock Decoupled weight decay regularization.
\newblock In \emph{7th International Conference on Learning Representations, {ICLR} 2019, New Orleans, LA, USA, May 6-9, 2019}. OpenReview.net, 2019.

\bibitem[Meng et~al.(2021)Meng, Zhao, Huang, and Zhou]{MagFace}
Qiang Meng, Shichao Zhao, Zhida Huang, and Feng Zhou.
\newblock Magface: {A} universal representation for face recognition and quality assessment.
\newblock In \emph{{IEEE} Conference on Computer Vision and Pattern Recognition, {CVPR} 2021, virtual, June 19-25, 2021}, pages 14225--14234. Computer Vision Foundation / {IEEE}, 2021.

\bibitem[Mittal et~al.(2012{\natexlab{a}})Mittal, Moorthy, and Bovik]{BRISQE_IQA}
Anish Mittal, Anush~Krishna Moorthy, and Alan~Conrad Bovik.
\newblock No-reference image quality assessment in the spatial domain.
\newblock \emph{{IEEE} Trans. Image Process.}, 21\penalty0 (12):\penalty0 4695--4708, 2012{\natexlab{a}}.

\bibitem[Mittal et~al.(2012{\natexlab{b}})Mittal, Soundararajan, and Bovik]{mittal2012making}
Anish Mittal, Rajiv Soundararajan, and Alan~C Bovik.
\newblock Making a ``completely blind'' image quality analyzer.
\newblock \emph{IEEE SPL}, 20\penalty0 (3):\penalty0 209--212, 2012{\natexlab{b}}.

\bibitem[Moschoglou et~al.(2017)Moschoglou, Papaioannou, Sagonas, Deng, Kotsia, and Zafeiriou]{agedb}
Stylianos Moschoglou, Athanasios Papaioannou, Christos Sagonas, Jiankang Deng, Irene Kotsia, and Stefanos Zafeiriou.
\newblock Agedb: The first manually collected, in-the-wild age database.
\newblock In \emph{2017 {IEEE} CVPRW, {CVPR} Workshops 2017, Honolulu, HI, USA, July 21-26, 2017}, pages 1997--2005. {IEEE} Computer Society, 2017.

\bibitem[Nguyen et~al.(2021)Nguyen, Bui, Duong, Bui, and Luu]{Nguyen_2021_CVPR}
Xuan-Bac Nguyen, Duc~Toan Bui, Chi~Nhan Duong, Tien~D. Bui, and Khoa Luu.
\newblock Clusformer: A transformer based clustering approach to unsupervised large-scale face and visual landmark recognition.
\newblock In \emph{Proceedings of the IEEE/CVF Conference on Computer Vision and Pattern Recognition (CVPR)}, pages 10847--10856, 2021.

\bibitem[Ou et~al.(2021)Ou, Chen, Zhang, Huang, Li, Li, Li, Cao, and Wang]{SDDFIQA}
Fu{-}Zhao Ou, Xingyu Chen, Ruixin Zhang, Yuge Huang, Shaoxin Li, Jilin Li, Yong Li, Liujuan Cao, and Yuan{-}Gen Wang.
\newblock {SDD-FIQA:} unsupervised face image quality assessment with similarity distribution distance.
\newblock In \emph{{IEEE} Conference on Computer Vision and Pattern Recognition, {CVPR} 2021, virtual, June 19-25, 2021}, pages 7670--7679. Computer Vision Foundation / {IEEE}, 2021.

\bibitem[Ou et~al.(2024)Ou, Li, Wang, and Kwong]{Ou_2024_CVPR}
Fu-Zhao Ou, Chongyi Li, Shiqi Wang, and Sam Kwong.
\newblock Clib-fiqa: Face image quality assessment with confidence calibration.
\newblock In \emph{Proceedings of the IEEE/CVF Conference on Computer Vision and Pattern Recognition (CVPR)}, pages 1694--1704, 2024.

\bibitem[Qin et~al.(2024)Qin, Wang, Deng, Wang, Chen, Hu, and Deng]{Swinface}
Lixiong Qin, Mei Wang, Chao Deng, Ke Wang, Xi Chen, Jiani Hu, and Weihong Deng.
\newblock Swinface: {A} multi-task transformer for face recognition, expression recognition, age estimation and attribute estimation.
\newblock \emph{{IEEE} Trans. Circuits Syst. Video Technol.}, 34\penalty0 (4):\penalty0 2223--2234, 2024.

\bibitem[Schlett et~al.(2022)Schlett, Rathgeb, Henniger, Galbally, Fi{\'{e}}rrez, and Busch]{DBLP:journals/csur/SchlettRHGFB22}
Torsten Schlett, Christian Rathgeb, Olaf Henniger, Javier Galbally, Julian Fi{\'{e}}rrez, and Christoph Busch.
\newblock Face image quality assessment: {A} literature survey.
\newblock \emph{{ACM} Comput. Surv.}, 54\penalty0 (10s):\penalty0 210:1--210:49, 2022.

\bibitem[Schlett et~al.(2024)Schlett, Rathgeb, Tapia, and Busch]{DBLP:journals/tbbis/SchlettRTB24}
Torsten Schlett, Christian Rathgeb, Juan~E. Tapia, and Christoph Busch.
\newblock Considerations on the evaluation of biometric quality assessment algorithms.
\newblock \emph{{IEEE} Trans. Biom. Behav. Identity Sci.}, 6\penalty0 (1):\penalty0 54--67, 2024.

\bibitem[Sengupta et~al.(2016)Sengupta, Chen, Castillo, Patel, Chellappa, and Jacobs]{cfp-fp}
Soumyadip Sengupta, Jun{-}Cheng Chen, Carlos~Domingo Castillo, Vishal~M. Patel, Rama Chellappa, and David~W. Jacobs.
\newblock Frontal to profile face verification in the wild.
\newblock In \emph{2016 {IEEE} Winter Conference on Applications of Computer Vision, {WACV} 2016, Lake Placid, NY, USA, March 7-10, 2016}, pages 1--9. {IEEE} Computer Society, 2016.

\bibitem[Shi and Jain(2019)]{PFE_FIQA}
Yichun Shi and Anil~K. Jain.
\newblock Probabilistic face embeddings.
\newblock In \emph{2019 {IEEE/CVF} International Conference on Computer Vision, {ICCV} 2019, Seoul, Korea (South), October 27 - November 2, 2019}, pages 6901--6910. {IEEE}, 2019.

\bibitem[Sun and Tzimiropoulos(2022)]{DBLP:conf/bmvc/SunT22}
Zhonglin Sun and Georgios Tzimiropoulos.
\newblock Part-based face recognition with vision transformers.
\newblock In \emph{33rd British Machine Vision Conference 2022, {BMVC} 2022, London, UK, November 21-24, 2022}, page 611. {BMVA} Press, 2022.

\bibitem[Terh{\"{o}}rst et~al.(2020)Terh{\"{o}}rst, Kolf, Damer, Kirchbuchner, and Kuijper]{SERFIQ}
Philipp Terh{\"{o}}rst, Jan~Niklas Kolf, Naser Damer, Florian Kirchbuchner, and Arjan Kuijper.
\newblock {SER-FIQ:} unsupervised estimation of face image quality based on stochastic embedding robustness.
\newblock In \emph{2020 {IEEE/CVF} Conference on Computer Vision and Pattern Recognition, {CVPR} 2020, Seattle, WA, USA, June 13-19, 2020}, pages 5650--5659. Computer Vision Foundation / {IEEE}, 2020.

\bibitem[Terh{\"o}rst et~al.(2020)Terh{\"o}rst, Kolf, Damer, Kirchbuchner, and Kuijper]{terhorst2020face}
Philipp Terh{\"o}rst, Jan~Niklas Kolf, Naser Damer, Florian Kirchbuchner, and Arjan Kuijper.
\newblock Face quality estimation and its correlation to demographic and non-demographic bias in face recognition.
\newblock In \emph{2020 IEEE Int.~Joint Conf.~on Biometrics (IJCB)}, pages 1--11. IEEE, 2020.

\bibitem[Wang et~al.(2018)Wang, Wang, Zhou, Ji, Gong, Zhou, Li, and Liu]{DBLP:conf/cvpr/WangWZJGZL018}
Hao Wang, Yitong Wang, Zheng Zhou, Xing Ji, Dihong Gong, Jingchao Zhou, Zhifeng Li, and Wei Liu.
\newblock Cosface: Large margin cosine loss for deep face recognition.
\newblock In \emph{2018 {IEEE} Conference on Computer Vision and Pattern Recognition, {CVPR} 2018, Salt Lake City, UT, USA, June 18-22, 2018}, pages 5265--5274. {IEEE} Computer Society, 2018.

\bibitem[Yang et~al.(2025)Yang, Grother, Ngan, Hanaoka, and Hom]{yang2025fate}
Joyce Yang, Patrick Grother, Mei Ngan, Kayee Hanaoka, and Austin Hom.
\newblock {Face Analysis Technology Evaluation (FATE) Part 11: Face Image Quality Vector Assessment – Specific Image Defect Detection}.
\newblock NIST Internal Report NIST IR 8485 DRAFT SUPPLEMENT, National Institute of Standards and Technology (NIST), Gaithersburg, MD, 2025.
\newblock Image Group, Information Access Division, Information Technology Laboratory. This publication is available free of charge.

\bibitem[You et~al.(2025)You, Li, Sun, Wei, Guo, Feng, and Ran]{You_2025_LVFace}
Jinghan You, Shanglin Li, Yuanrui Sun, Jiangchuan Wei, Mingyu Guo, Chao Feng, and Jiao Ran.
\newblock Lvface: Progressive cluster optimization for large vision models in face recognition, 2025.

\bibitem[Zhang et~al.(2023)Zhang, Li, Liu, Zhang, Su, Zhu, Ni, and Shum]{DBLP:conf/iclr/0097LL000NS23}
Hao Zhang, Feng Li, Shilong Liu, Lei Zhang, Hang Su, Jun Zhu, Lionel~M. Ni, and Heung{-}Yeung Shum.
\newblock {DINO:} {DETR} with improved denoising anchor boxes for end-to-end object detection.
\newblock In \emph{The Eleventh International Conference on Learning Representations, {ICLR} 2023, Kigali, Rwanda, May 1-5, 2023}. OpenReview.net, 2023.

\bibitem[Zhang et~al.(2016)Zhang, Zhang, Li, and Qiao]{zhang2016joint}
Kaipeng Zhang, Zhanpeng Zhang, Zhifeng Li, and Yu Qiao.
\newblock Joint face detection and alignment using multitask cascaded convolutional networks.
\newblock \emph{{IEEE} Signal Process. Lett.}, 23\penalty0 (10):\penalty0 1499--1503, 2016.

\bibitem[Zheng and Deng(2018)]{CPLFWTech}
T. Zheng and W. Deng.
\newblock Cross-pose lfw: A database for studying cross-pose face recognition in unconstrained environments.
\newblock Technical Report 18-01, Beijing University of Posts and Telecommunications, 2018.

\bibitem[Zheng et~al.(2017)Zheng, Deng, and Hu]{CALFW}
Tianyue Zheng, Weihong Deng, and Jiani Hu.
\newblock Cross-age {LFW:} {A} database for studying cross-age face recognition in unconstrained environments.
\newblock \emph{CoRR}, abs/1708.08197, 2017.

\bibitem[Zhong and Deng(2021)]{Zhong_2021_FT}
Yaoyao Zhong and Weihong Deng.
\newblock Face transformer for recognition.
\newblock \emph{arXiv preprint arXiv:2103.14803}, 2021.

\bibitem[Žiga Babnik et~al.(2022)Žiga Babnik, Peer, and Štruc]{FaceQAN}
Žiga Babnik, Peter Peer, and Vitomir Štruc.
\newblock {FaceQAN}: Face image quality assessment through adversarial noise exploration.
\newblock In \emph{26th International Conference on Pattern Recognition (ICPR 2022)}, pages 748--754, Montreal, Canada, 2022. IEEE.

\end{thebibliography}
}

\end{document}